\documentclass[11pt]{article}

\usepackage[preprint]{acl}

\usepackage{times}
\usepackage{latexsym}

\usepackage[T1]{fontenc}
\usepackage[utf8]{inputenc}

\usepackage{microtype}

\usepackage{inconsolata}

\usepackage{graphicx}
\graphicspath{{images/}}
\DeclareGraphicsExtensions{.pdf,.png,.jpg,.jpeg}

\usepackage{subcaption}
\usepackage{amsmath}
\usepackage{amssymb}
\usepackage{amsthm}
\usepackage{mathrsfs}
\usepackage{multirow}
\usepackage{cleveref}
\usepackage[x11names]{xcolor}
\usepackage{colortbl}
\usepackage{enumitem}
\usepackage{booktabs}
\usepackage{tabularx}
\usepackage{array}
\usepackage{float}

\newcommand{\seg}{s}
\newcommand{\segSet}{S}
\newcommand{\SETsegSet}{\mathcal{S}}
\newcommand{\segi}[1]{s_{#1}}
\newcommand{\doc}{D}
\newcommand{\doci}[1]{D_{#1}}
\newcommand{\segiDocj}[2]{\segi{#1}^{\doci{#2}}}

\newcommand{\embedModel}{F_{\theta}}
\newcommand{\embedModelt}[1]{\embedModel({#1})}

\newcommand{\poolingToken}{\texttt{<s>}}
\newcommand{\eosToken}{\texttt{</s>}}
\newcommand{\embedText}[1]{\mathbf{e}_{#1}}
\newcommand{\embedSegiDocjSTD}[2]{\embedText{\segiDocj{#1}{#2}}^{iso}} %Standalone segment embedding (embedding of segment i within document j)
\newcommand{\embedDoci}[1]{\embedText{\doci{#1}}} %Document embedding
\newcommand{\embedSegiDocjCTX}[2]{\embedText{\segiDocj{#1}{#2}}^{ctx}} %Contextualized segment embedding (embedding of segment i within document j)

\newcommand{\lang}{\mathcal{L}}

\newcommand{\langConfig}{\mathbf{L}}
\newcommand{\langFirst}{\mathcal{L}_{lead}}
\newcommand{\langRemainder}{\mathcal{L}_{later}}

\newcommand{\similarityOneij}[2]{\sigma_{#1,#2}}

\newcommand{\similarityTwoij}[2]{\tau_{#1,#2}}

\newcommand{\olsBetai}[1]{\beta_{#1}}
\newcommand{\olsIntercept}{Intercept}
\newcommand{\olsPi}[1]{#1~vs.~1}

\newcommand{\numValidTokens}{L}

\newcommand{\basketSize}{\mathfrak{B}}

\newcommand{\layersCal}{\mathfrak{L}^C}

\newcommand{\numLayers}{\mathscr{L}}

\newcommand{\alphabet}{\Sigma}
\newcommand{\alphabetKleene}{\Sigma^{*}}

\DeclareRobustCommand{\bc}[1][1.0]{\ensuremath{\scalebox{#1}{$\bullet$}}}
\newcommand{\bclarge}{\bc[0.8]}
\DeclareRobustCommand{\os}[1][1.0]{\ensuremath{\scalebox{#1}{$\square$}}}
\newcommand{\osmedium}{\os[0.5]}
\DeclareRobustCommand{\od}[1][1.0]{\ensuremath{\scalebox{#1}{$\lozenge$}}}
\newcommand{\odmedium}{\od[0.5]}

\newcommand{\snn}{$^{\bclarge}$}
\newcommand{\son}{$^{\osmedium}$} %p<0.05
\newcommand{\stw}{$^{\odmedium}$} %p<0.01
\newcommand{\sth}{} %p<0.001

\newcommand{\cEN}[1]{{\color{PeachPuff4}{#1}}}
\newcommand{\cENbold}[1]{\textbf{\color{PeachPuff4}{#1}}}

\newcommand{\cDE}[1]{{\color{PaleGreen4}{#1}}}
\newcommand{\cDEbold}[1]{\textbf{\color{PaleGreen4}{#1}}}

\title{Measuring and Mitigating Positional Bias of Long Text Embedding Models}

\title{Information Representation Fairness in Long-Document Embeddings: The Peculiar Interaction of Positional and Language Bias}

\author{
    {\bf Elias Schuhmacher\thanks{Equal contribution.} \hspace{2mm}}
    {\bf Andrianos Michail\textsuperscript{\thefootnote} \hspace{2mm}}
    {\bf Juri Opitz \hspace{2mm}} \\
    {\bf Rico Sennrich \hspace{2mm}} 
    {\bf Simon Clematide} \\
    Department of Computational Linguistics \\
    University of Zurich \\
    \texttt{<firstname>.<lastname>@uzh.ch}}

\usepackage{subfiles}

\begin{document}
\maketitle
\begin{abstract}

To be discoverable in an embedding-based search process, each part of a document should be reflected in its embedding representation. To quantify any potential reflection biases, we introduce a permutation-based evaluation framework. With this, we observe that state-of-the-art embedding models exhibit systematic positional and language biases when documents are longer and consist of multiple segments. Specifically, early segments and segments in higher-resource languages like English are over-represented, while later segments and segments in lower-resource languages are marginalized. In our further analysis, we find that the positional bias stems from front-loaded attention distributions in pooling-token embeddings, where early tokens receive more attention. To mitigate this issue, we introduce an inference-time attention calibration method that redistributes attention more evenly across document positions, increasing discoverabiltiy of later segments. Our evaluation framework and attention calibration is available at \href{https://github.com/impresso/fair-sentence-transformers}{github.com/impresso/fair-sentence-transformers}
\end{abstract}

\section{Introduction}
Text embedding models serve as the backbone of search engines and retrieval modules in Retrieval Augmented Generation (RAG) and agentic systems. These models map documents and queries into a shared vector space where cosine similarity determines discoverability—any distortion during embedding directly affects which information can be retrieved. Previous work has shown that models exhibit a pronounced \emph{positional bias}—prioritizing information based on where it appears \citep[e.g.,][]{zhu_longembed_2024,zeng-etal-2025-empirical,fayyaz-etal-2025-collapse,ognawala_long-context_2025,liu_lost_2024,Mohtashami-Jaggi-random-access-2023,lee2025quantifyingpositionalbiasestext,coelho_dwell_2024,modarressi2025nolima}.

Longer documents often place distinct information in different sections—legal documents put definitions in appendices, technical manuals include troubleshooting after introductory content. When embedding models prioritize early content, key information in later positions becomes invisible to semantic search. Consider newspaper pages with multiple independent articles, sometimes in different languages. If the model prioritizes early content, later articles become undiscoverable. Structural conventions and placement determine which information is matched, often over relevance.

\begin{figure}[t] 
\centering
\includegraphics[width=\linewidth, trim=5 2 2 2, clip]{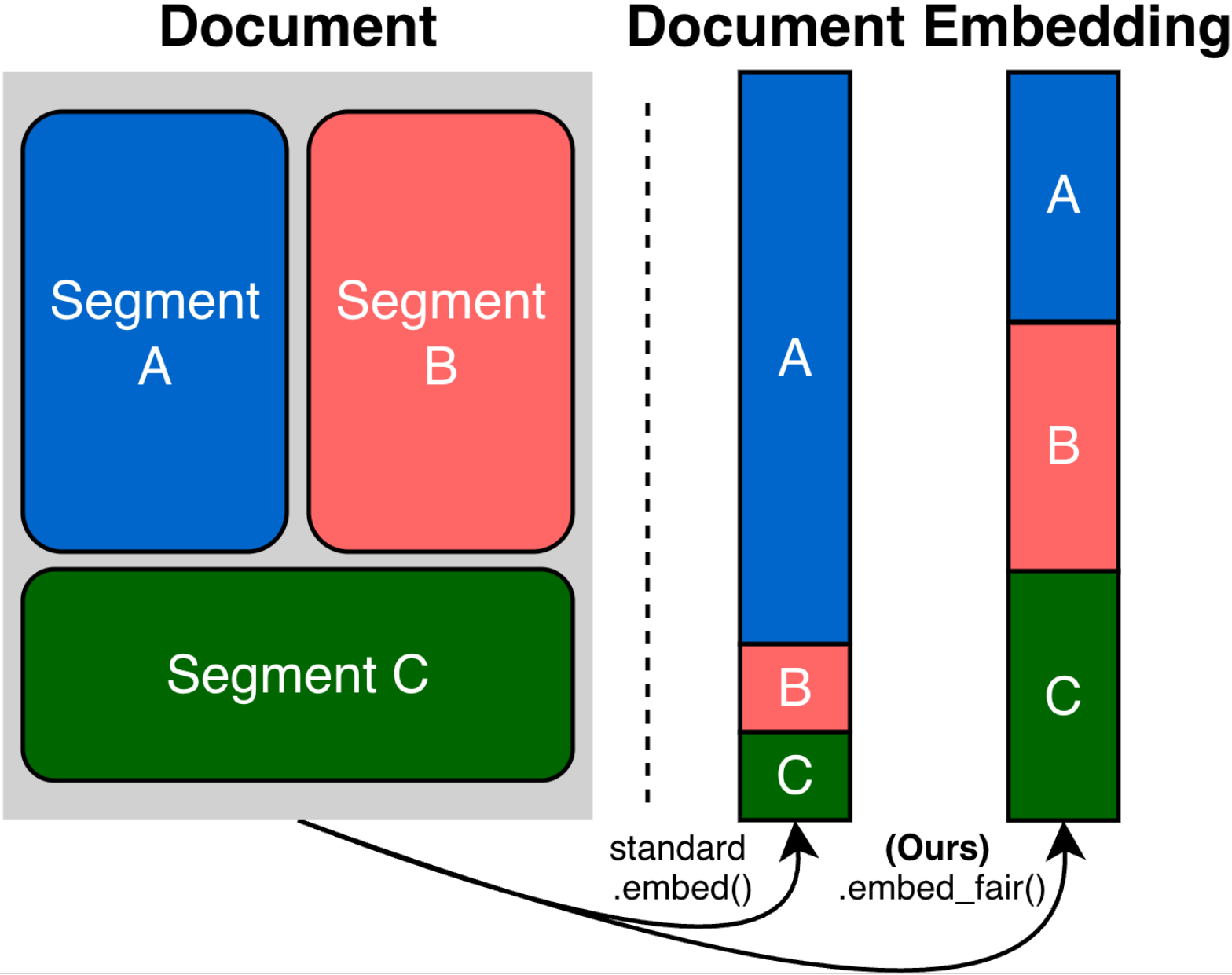} 
\caption{We demonstrate that standard document encoders generate skewed embeddings that underrepresent later segments when processing multi-segment documents. We show that an inference-time attention calibration method yields embeddings that are fairer regardless of the relative order of the segment.}
\label{fig:sample_frontpage}
\end{figure}

Figure~\ref{fig:sample_frontpage} illustrates this problem: due to positional bias, the representation of a multi-segment document is dominated by the first segment, making information from later segments invisible.

\begin{figure}[t] 
\centering
\includegraphics[width=\linewidth, trim=32 34 32 16, clip]{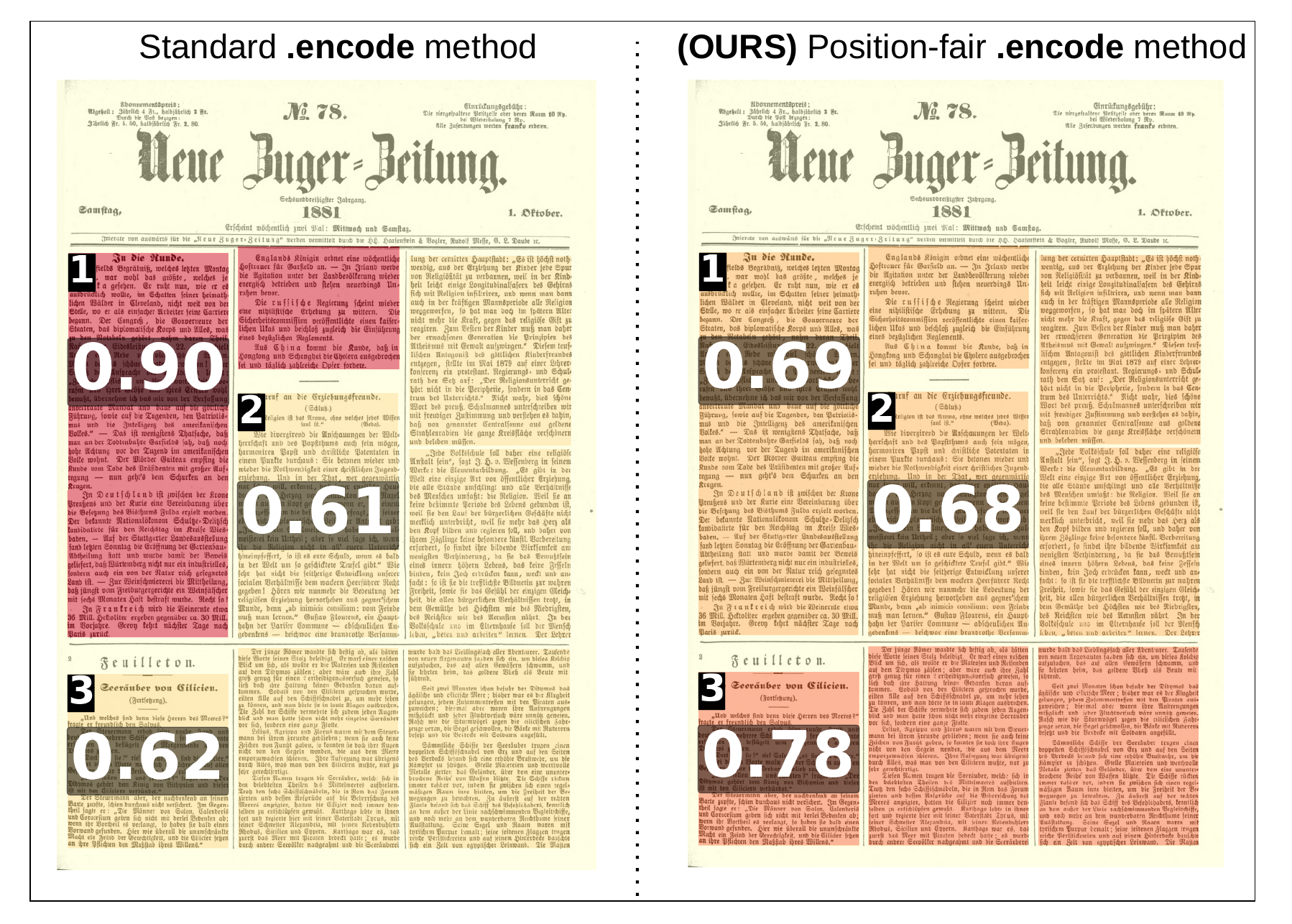} 
\caption{A case study that visualizes how positional bias can lead to unfair content representation (left) and shows the effect of applying our proposed mitigation (right). Concretely, the image shows the similarities between the page embedding and the standalone embeddings of three articles. Standard encoding (left) over-represents the first article. Attention calibration (right) distributes similarity more evenly across articles.}
\label{fig:sample_newspaper}
\end{figure}

We formalize this as a threat to \emph{Information Representation Fairness}: when a document composed of multiple segments is embedded, each segment should contribute equally to the resulting vector representation, regardless of its position or language. Our analysis focuses on a general search scenario in which a user searches for document-level information using one of its segments. In this scenario, a fair document embedding should be equally similar to all of its constituent segments, regardless of segment position or source language. Our setting most closely resembles page-level newspaper embeddings, where distinct articles are concatenated in arbitrary order—making discoverability largely determined by whichever article appears first. Figure~\ref{fig:sample_newspaper} illustrates this issue on a real document from an historic newspaper, showing the practical impact of our calibration method.

As first steps toward fairer long-document embedding models, our contributions are:
\begin{itemize}
    \item An evaluation framework that uses a comparable corpus to quantify positional biases in long-context embedding models.
    \item A mixed-language extension that reveals source language bias as an additional dimension of unfairness.
    \item An analysis of the attention patterns of embedding models that use the \poolingToken-token as the pooling token, identifying attention-based causes of positional bias.
    \item An inference-time attention calibration method that mitigates the positional bias and produces fairer embeddings.
\end{itemize}
By providing this methodology to detect, quantify, and explain prevalent biases, we advance the understanding of long-document embeddings, a pertinent interpretability challenge \citep{opitz-etal-2025-interpretable}.

\section{Related Work}
\subsection*{Positional Bias in Long-Context Models}
Early evidence of long-context limitations of transformer-based models was provided by \citet{Mohtashami-Jaggi-random-access-2023}, who introduced a passkey retrieval challenge to test decoder language models and showed that these models often fail to retrieve the correct key when it is buried deep in the input. Similarly, \citet{liu_lost_2024} observe a ``lost in the middle'' phenomenon stating that decoder models struggle to use information placed in the middle of their input context as opposed to information placed at the start or end.
\citet{lee2025quantifyingpositionalbiasestext} showed that embeddings of encoder models are more strongly influenced by perturbations at the start of a text than by identical perturbations in the middle or at the end. \citet{zeng-etal-2025-empirical} empirically demonstrated that, when performing question answering retrieval, embedding models exhibit higher performance degradation when the related content is placed later within the context.

\subsection*{Attention Calibration in Decoder-Based Language Models}
\citet{hsieh_found_2024} proposed a method to counteract the U-shaped positional bias in decoder-based language models, attributing the “lost in the middle” phenomenon \citep{liu_lost_2024} to a corresponding U-shaped pattern in decoder self-attention. This method estimates positional bias by performing one inference per position while shifting a fixed dummy segment across all (e.g., 20) positions, thereby obtaining a baseline attention distribution that reflects positional bias. The estimated bias is then subtracted from the observed segment attention to produce calibrated attention, which is used during a “positionally fair” inference.

\section{Methods}

We introduce the notation of \emph{segments} and \emph{documents} used in this work and define the two concepts: \emph{positional fairness} and \emph{information retention}.

\subsection*{Notation}
Let $\alphabet$ be a finite vocabulary (alphabet) of tokens, and let $\alphabetKleene$ denote its Kleene closure–the set of all finite token sequences over $\alphabet$. We define a \emph{segment} $\seg \in \alphabetKleene$: a logically coherent unit of text of arbitrary length. Each segment is treated as an indivisible unit of information. Further, we define a \emph{segment set} $\segSet = \{\seg_1,\ldots,\seg_n\}$: an unordered collection of $n$ segments, with $\seg_i \neq \seg_h$ for $i\neq h$. Finally, we define a \emph{document} $\doc \in \alphabetKleene$: a concatenation of multiple segments. Given a segment set $\segSet$, we construct $n!$ distinct documents by permuting segment order. We employ this permutation setup to eliminate the segment content as a confounding factor in our analysis. Given a segment set $\segSet$ of size $n \geq1$ and a specific ordering $j$ out of all $n!$ distinct orderings, document $\doci{j}$ is the concatenation (with whitespace) of the $n$ segments in $\segSet$ following this ordering. We denote the segment at position $i$ within document $\doci{j}$ as $\segiDocj{i}{j}$, and use {1-indexed} positions.

\begin{figure*}[t]
  \centering
\includegraphics[width=0.967\textwidth]{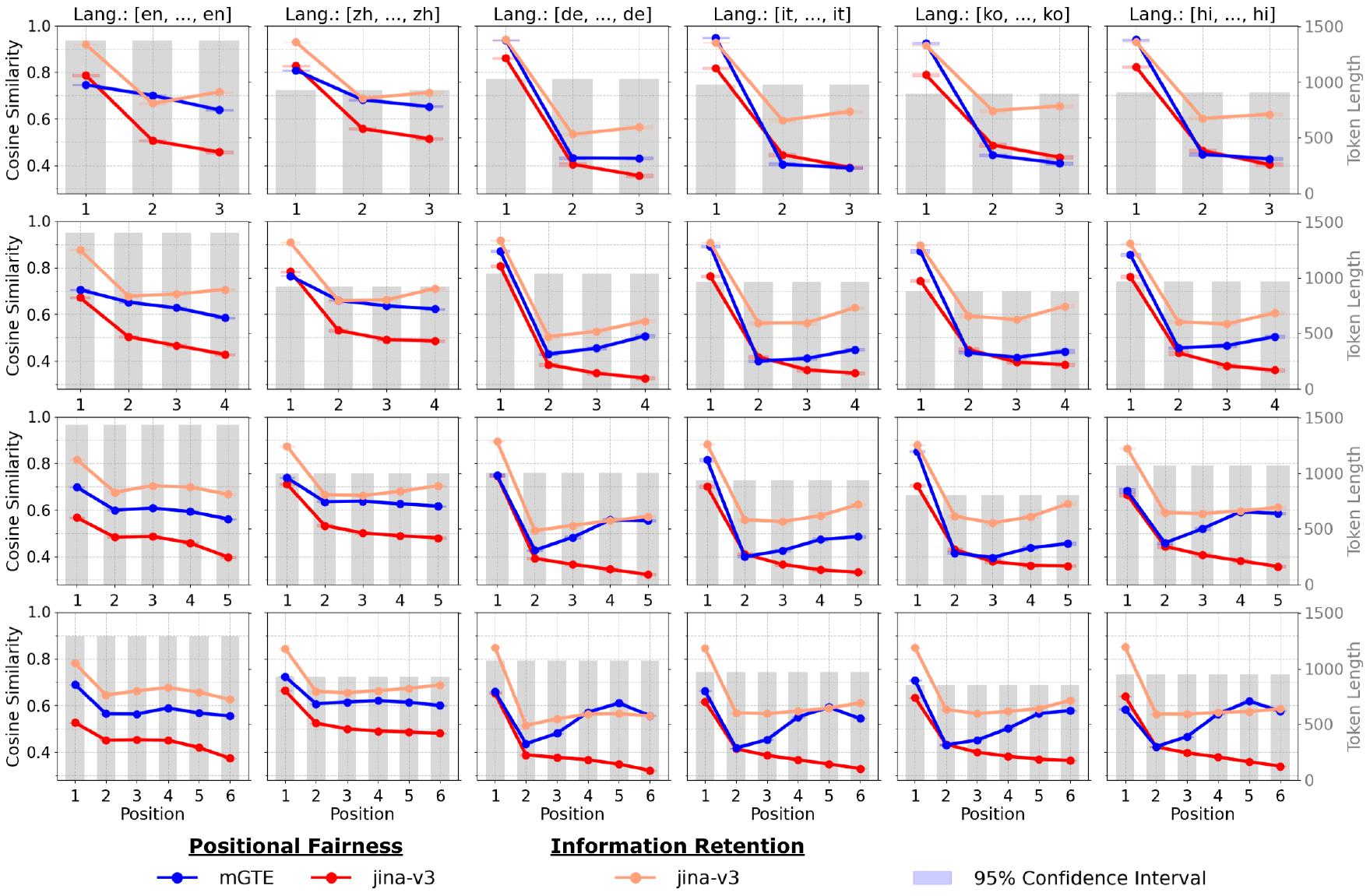}
  \caption{Monolingual experiment instances $(n, \langConfig)$, where $n$ varies across rows, and $\langConfig$ varies across columns. Left y-axes show (i) average representation in the global document embedding (\textcolor{Blue1}{mGTE} and \textcolor{Red2}{jina-v3}), and (ii) average information retention (\textcolor{LightSalmon1}{jina-v3}) per segment position. (\textcolor{Snow4}{Gray bars}) show average token length per segment position.}
  \label{fig:EF_IR_mono}
\end{figure*}

We consider \emph{text embedding models} with $\numLayers$ transformer layers, following the functional form: $\embedModel:\alphabetKleene \to \mathbb{R}^{d}$.

We write $\embedSegiDocjSTD{i}{j} = \embedModelt{\segiDocj{i}{j}}$ to denote the embedding of segment $i$ from document $\doci{j}$. Note that this is the embedding of the segment \emph{isolated} on its own as standalone text, i.e., without contextual information from other segments in $\doci{j}$. Similarly, $\embedDoci{j} = \embedModelt{\doci{j}}$ the embedding of a document $\doci{j}$.

For mean-pooled models, we write $\embedSegiDocjCTX{i}{j}$ to denote the \emph{contextualized} segment embedding (late chunking; \citep{günther2025latechunkingcontextualchunk}) of the segment at position $i$ within document $\doci{j}$:
\begin{equation}
  \embedSegiDocjCTX{i}{j} = \frac{1}{|P(\segiDocj{i}{j})|} \sum_{p \in P(\segiDocj{i}{j})} \mathbf{h}_p^{(\numLayers)} \nonumber,
\end{equation}
with $\mathbf{h}_p^{(\numLayers)}$ the contextual token representation at the final layer $\numLayers$ of embedding model $\embedModel$ for the token at position index $p$, $P(\segiDocj{i}{j})$ the set of token index positions belonging to segment $\segiDocj{i}{j}$ within document $\doci{j}$, and $|P(\segiDocj{i}{j})|$ the number of tokens in segment $\segiDocj{i}{j}$. As segment-wise mean pooling requires meaningful token-level representations, this operation is only applicable for embedding models preserving token-level semantics during fine-tuning, such as mean-pooled models.

\begin{figure*}[t]
  \centering
\includegraphics[width=0.967\textwidth]{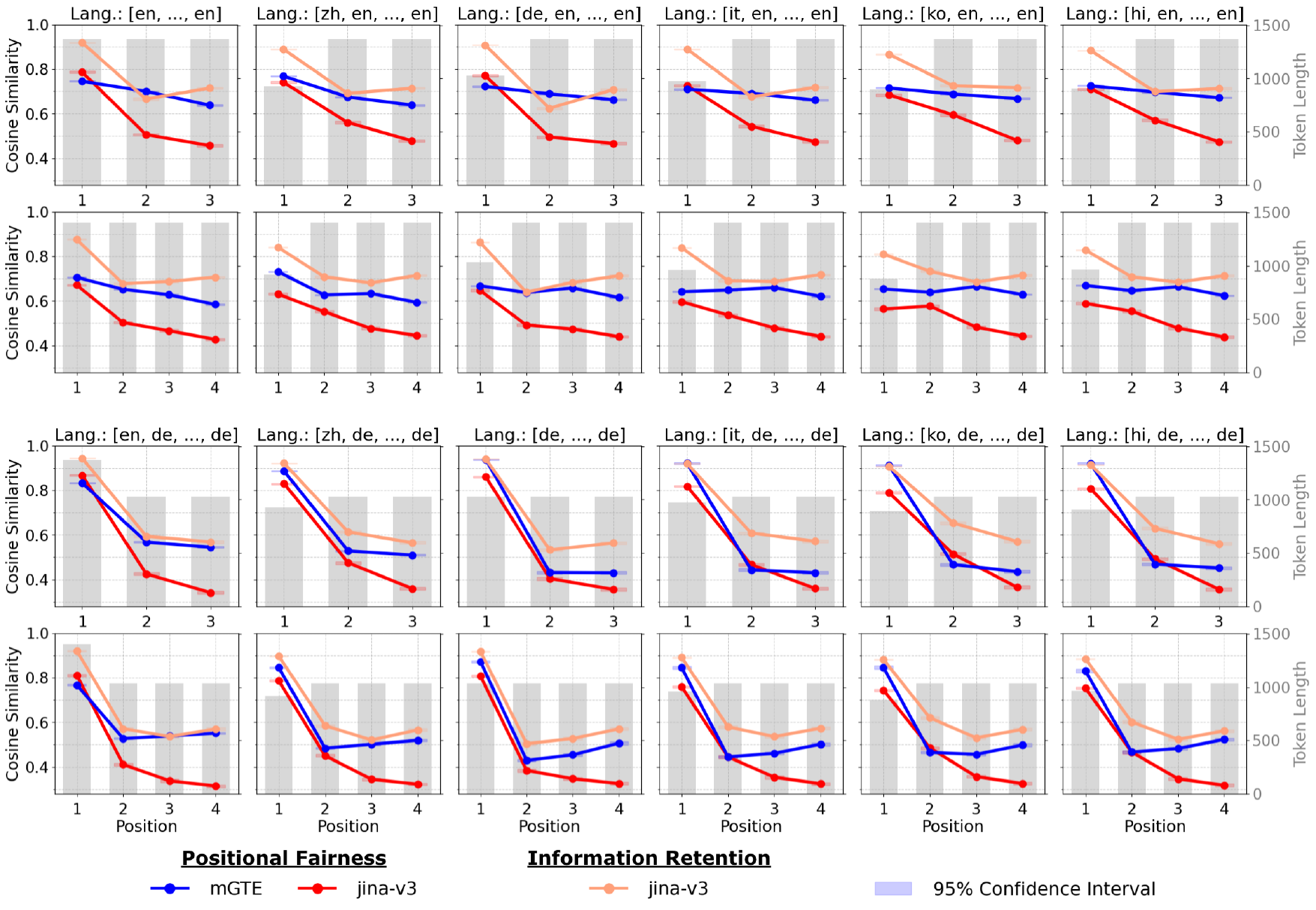}
  \caption{Preferential treatment of English segments: later segments in $\langRemainder=en$ (top 2 rows) are better represented in the global document embedding (\textcolor{Blue1}{mGTE} and \textcolor{Red2}{jina-v3}) and exhibit higher information retention (\textcolor{LightSalmon1}{jina-v3}) than later segments in $\langRemainder=de$ (bottom 2 rows).}
  \label{fig:EF_IR_multi_EN_DE_direct_comparison}
\end{figure*}

\subsection*{Construction of Segments and Documents}
We construct a multilingual comparable corpus based on topic-aligned Wikipedia articles in six languages.  We include high-resource (English, Chinese, German, Italian) and lower-resource (Korean, Hindi) languages. To improve cross-lingual semantic alignment, we apply heuristic length-based filtering, retaining articles whose lengths are within ±70\% of the English article length, measured in XLM-R subtokens. On average, articles are shorter than English by 31\% ($zh$), 25\% ($de$), 30\% ($it$), 37\% ($ko$), and 29\% ($hi$). A single sample is shown in \Cref{fig:wiki-comparable-sample} in the Appendix.

Each segment $\seg$ is a Wikipedia article between 1,000 and 2,000 tokens. Each document $\doc$ consists of $n$ randomly drawn segments, with a maximum document length of 8,192 tokens. We set $n\in\{3,4,5,6\}$. Documents follow a language configuration $\langConfig = (\lang_1, \ldots, \lang_n)$, where $\lang_i \in \{en,zh,de,it,ko,hi\}$ specifies the language at position $i$. We construct \textbf{monolingual documents} with $\langConfig = (\lang, \ldots, \lang)$, and \textbf{mixed-language documents} with $\langConfig = (\langFirst, \langRemainder, \ldots, \langRemainder)$ and $\langFirst \neq \langRemainder$. Each combination of $(n, \langConfig)$ defines an experiment instance. Owing to the factorial number of segment permutations, we construct 1,002 documents for $n{=}3$, 1,008 for $n{=}4$, 1,080 for $n{=}5$, and 10,080 for $n{=}6$.

\subsection*{Examined Embedding Models}
We analyze the information representation fairness of two state-of-the-art multilingual encoder-based embedding models that employ different pooling strategies to produce document embeddings. 
\textbf{mGTE}\footnote{\href{https://huggingface.co/Alibaba-NLP/gte-multilingual-base}{Alibaba-NLP/gte-multilingual-base}} represents a document using the  \textbf{start-of-sequence token} $\poolingToken$  embedding  \citep{zhang_mgte_2024}.
\textbf{jina-v3}\footnote{\href{https://huggingface.co/jinaai/jina-embeddings-v3}{jinaai/jina-embeddings-v3}}
uses \textbf{mean pooling} over contextualized token representations from a LoRA adapter \citep{sturua_jina-embeddings-v3_2024}. We use the text-matching LoRA adapter.

Both models (i) support long contexts of up to 8,192 tokens, (ii) cover all six languages considered in this study, and (iii) perform well on the Multilingual MTEB benchmark for long-context ($\ge8,192$ tokens)  \citep{enevoldsen2025mmtebmassivemultilingualtext}.

\subsection*{Positional Fairness}
Let $Q_{\segSet}$ denote a data-generating process of $n$-sized segment sets $\segSet$, each containing $n$ Wikipedia articles. We write $\operatorname{supp}(Q_{\segSet})$ for the set of segment sets that occur with positive probability under $Q_{\segSet}$. For a fixed $\segSet$, let $J \mid \segSet \sim \mathrm{Unif}\{1,\ldots,n!\}$ be a specific ordering sampled uniformly from all possible $n!$ permutations, and let $\doci{J}$ be the corresponding document. Let
$\similarityOneij{i}{J} \;=\; \cos\!\big(\embedDoci{J},\, \embedSegiDocjSTD{i}{J}\big)$ be the cosine similarity between the document embedding and the standalone embedding of the segment placed at position $i$ in $\doci{J}$.

An embedding model is \mbox{\emph{fair}} if
$
    \mathbb{E}\!\left[\,\similarityOneij{1}{J} \;\middle|\; \segSet \right]
    \;=\; \cdots \;=\;
    \mathbb{E}\!\left[\,\similarityOneij{n}{J} \;\middle|\; \segSet \right]
    \quad \forall\, \segSet \in \operatorname{supp}(Q_{\segSet}),
$
i.e., the document embedding represents the semantic content of all constituent segments equally well \emph{regardless of their position} in the document.

We fit an ordinary least squares regression (OLS; \citet{gelman_regression_2021_ols}) with categorical indicators for positions to estimate deviations from positional fairness. Let $\SETsegSet$ be the finite collection of segment sets in our dataset. For each $\segSet \in \SETsegSet$, for each permutation $j \in \{1,\ldots,n!\}$ of that set, and for each position $i \in \{1,\ldots,n\}$, we observe the similarity $\similarityOneij{i}{j}$. We pool all observations across $\segSet \in \SETsegSet$ and estimate
$
  \similarityOneij{i}{j}
  \,=\, \olsBetai{0}
  \,+\, \sum_{p=2}^{n} \olsBetai{p}\, \mathbb{I}\{i=p\}
  \,+\, \varepsilon_{i,j,\segSet} \nonumber,
$
where $\olsBetai{p}$ captures the difference of position $p$ relative to the baseline (similarity at position~1). Standard errors are clustered at the \emph{segment-set} level, i.e., $\varepsilon_{i,j,\segSet}$ may be arbitrarily correlated within a given $\segSet$ but is assumed uncorrelated across different segment sets. The null hypothesis of \emph{positional fairness} is
$H_0^{(p)}:\; \olsBetai{p} = 0 \quad (p=2,\ldots,n)$.%

\subsection*{Information Retention}
Let $\similarityTwoij{i}{J} \;=\; \cos\!\big(\embedSegiDocjSTD{i}{J},\, \embedSegiDocjCTX{i}{J}\big)$ be the cosine similarity between the standalone embedding and the \emph{contextualized} embedding of the same segment if placed at position $i$ in $\doci{J}$. This measure quantifies how strongly the semantic information of a segment changes when contextualized inside a longer, multi-segment document.

An embedding model exhibits \emph{no position-dependent information retention} if
$
    \mathbb{E}\!\left[\,\similarityTwoij{1}{J} \;\middle|\; \segSet \right]
    \;=\; \cdots \;=\;
    \mathbb{E}\!\left[\,\similarityTwoij{n}{J} \;\middle|\; \segSet \right]
    \quad \forall\, \segSet \in \operatorname{supp}(Q_{\segSet}),
$
i.e., the semantics of any segment is equally well preserved during contextualization \emph{regardless of its position} within a document. We estimate
$
  \similarityTwoij{i}{j}
  \,=\, \olsBetai{0}^{(\tau)}
  \,+\, \sum_{p=2}^{n} \olsBetai{p}^{(\tau)}\, \mathbb{I}\{i=p\}
  \,+\, \varepsilon^{(\tau)}_{i,j,\segSet} \nonumber,
$
with the null hypothesis $H_0^{(p)}:\; \olsBetai{p}^{(\tau)} = 0 \quad (p=2,\ldots,n)$.

\section{Analysis}

\subsection{Monolingual Documents}

\subsubsection*{Positional Fairness}

\Cref{fig:EF_IR_mono} depicts the segment-representation profiles of the $n$×$\langConfig$ (4×6) combinations of monolingual experiment instances (shown in \textcolor{Blue1}{blue (mGTE)} and \textcolor{Red2}{dark red (jina-v3)}). We observe a consistent positional bias: any segment—if placed at the first position of a document—is captured substantially better in the global document embedding than other segments. This results in L-shaped segment-representation profiles. The strength of this positional bias varies by language: the difference in the document representation between the first-positioned segment and any later segment is markedly smaller for English and Chinese.

The OLS estimates strongly reject the null hypotheses $H_0^{(p)}: \olsBetai{p}=0$ for most $p$ in all experiment instances. The largest negative coefficients typically occur at $p{=}2$ and $p{=}3$, indicating that the semantic content of second- and third-positioned segments is most underrepresented in the global document representation. Detailed OLS estimates are shown in \Cref{tab:ols-coef-mono-mgte,tab:ols-coef-mono-jina} in the Appendix. The magnitude of the positional effects attenuates for later positions.

% \footnote{For mGTE, baseline segment-representations of $n{=}3$ experiment instances for \textit{(\cEN{en}, zh, \cDE{de}, it, ko, hi)} are \textit{(\cEN{0.75}\sth, 0.81\sth, \cDE{0.94}\sth, 0.95\sth, 0.93\sth, 0.94\sth)} at position~1 ($\olsBetai{0}$); differences to baseline position~1 at position~2 ($\olsBetai{2}$) are \textit{(\cEN{-0.05}\sth, -0.13\sth, \cDEbold{-0.51}\sth, -0.54\sth, -0.48\sth, -0.49\sth)}. Detailed OLS estimates are shown in \cref{tab:ols-coef-mono-mgte,tab:ols-coef-mono-jina} in the Appendix.}

\subsubsection*{Information Retention}
\Cref{fig:EF_IR_mono} shows the segment-retention profiles of the 4×6 monolingual experiment instances (shown in \textcolor{LightSalmon1}{light red (jina-v3)}). We identify a clear pattern: first-positioned segments retain most of their semantic content, whereas later-positioned segments show progressively greater divergence between their standalone and contextualized information. This results in L-shaped segment-retention profiles. The magnitude of this positional effect varies by language: English and Chinese segments show greater resilience to semantic distortion. We strongly reject the null hypotheses $H_0^{(p)}: \olsBetai{p}^{(\tau)}=0$ for all $p$ in all experiment instances. Detailed OLS estimates are shown in \Cref{tab:ols-coef-mono-jina-exp2} in the Appendix.

\begin{figure*}[t]
  \centering
\includegraphics[width=\textwidth, trim=28 5.5 28 5.5, clip]{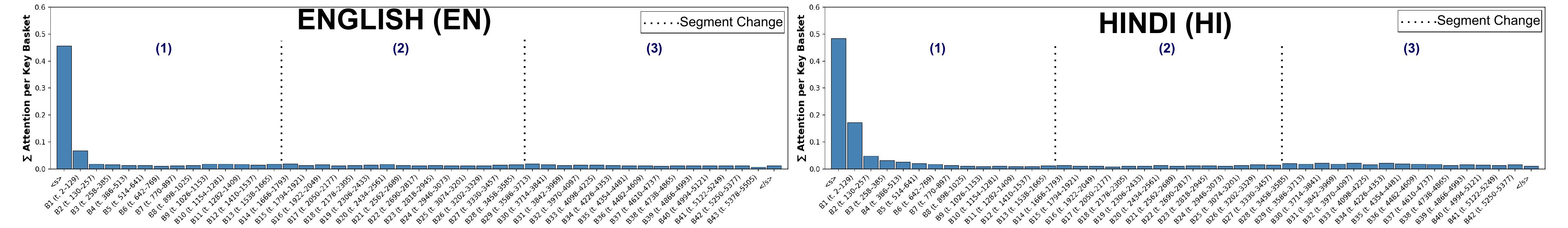}
\caption{Front-loaded self-attention distribution of the $\poolingToken$-query token over key baskets (basket size $\basketSize{=}128$) in English (left) and Hindi (right) documents ($n{=}3$). Average of the last six transformer layers.}
\label{fig:attention_combined}
\end{figure*}

% \footnote{Baseline segment-retentions of $n{=}3$ experiment instances for \textit{(\cEN{en}, zh, \cDE{de}, it, ko, hi)} are \textit{(\cEN{0.92}\sth, 0.93\sth, \cDE{0.94}\sth, 0.93\sth, 0.92\sth, 0.93\sth)} at position~1 ($\olsBetai{0}^{(\tau)}$); differences to baseline position~1 at position~2 ($\olsBetai{2}^{(\tau)}$) are \textit{(\cEN{-0.25}\sth, -0.24\sth, \cDEbold{-0.41}\sth, -0.34\sth, -0.28\sth, -0.33\sth)}. Detailed OLS estimates are shown in \Cref{tab:ols-coef-mono-jina-exp2} in the Appendix.}

\subsection{Mixed-Language Documents}

\subsubsection*{Positional Fairness}
For mixed-language documents, positional bias persists but is significantly influenced by language-specific effects. Most notably, English (and, for mGTE, also Chinese) segments are well represented in the global document embedding, regardless of the positional placement in the document. If placed at later positions, this language preference counteracts positional bias. \Cref{fig:EF_IR_multi_EN_DE_direct_comparison} illustrates this preferential treatment of English segments compared to German segments: experiment instances with $\langRemainder=en$ (top 2 rows) exhibit flatter segment-representation profiles (shown in \textcolor{Blue1}{blue (mGTE)} and \textcolor{Red2}{dark red (jina-v3)}) compared to $\langRemainder=de$ (bottom 2 rows). Detailed results of all language configurations are shown in \Cref{fig:EF_IR_multi_EN,fig:EF_IR_multi_DE,fig:EF_IR_multi_HI,fig:EF_IR_multi_ZH} in the Appendix. The OLS estimates confirm these results, detailed in \Cref{tab:ols-coef-multi-en-mgte,tab:ols-coef-multi-zh-mgte,tab:ols-coef-multi-de-mgte,tab:ols-coef-multi-hi-mgte,tab:ols-coef-multi-en-jina,tab:ols-coef-multi-zh-jina,tab:ols-coef-multi-de-jina,tab:ols-coef-multi-hi-jina} in the Appendix.
% \footnote{For mGTE, baseline segment-representations of $n{=}3$ experiment instances for \textit{([it,\cEN{en},\cEN{...},\cEN{en}], [it,\cDE{de},\cDE{...},\cDE{de}])} are \textit{(\cEN{0.71}\sth, \cDE{0.92}\sth)} at position~1 ($\olsBetai{0}$); differences to baseline position~1 at position~2 ($\olsBetai{2}$) are \textit{(\cEN{-0.02}\sth, \cDEbold{-0.48}\sth)}. Detailed OLS estimates are shown in \Cref{tab:ols-coef-multi-en-mgte,tab:ols-coef-multi-zh-mgte,tab:ols-coef-multi-de-mgte,tab:ols-coef-multi-hi-mgte,tab:ols-coef-multi-en-jina,tab:ols-coef-multi-zh-jina,tab:ols-coef-multi-de-jina,tab:ols-coef-multi-hi-jina} in the Appendix.}

\subsubsection*{Information Retention}
Positional effects remain in mixed-language documents, but similar to positional fairness, they exhibit language-specific effects. Specifically, English segments show greater resilience to interference from surrounding multilingual segments compared to other languages. \Cref{fig:EF_IR_multi_EN_DE_direct_comparison} shows these language-specific effects: experiment instances with $\langRemainder=en$ (top 2 rows) exhibit flatter information-retention profiles (shown in \textcolor{LightSalmon1}{light red (jina-v3)}) than the corresponding instances with $\langRemainder=de$ (bottom 2 rows). In the Appendix, detailed results of all language configurations are shown in \Cref{fig:EF_IR_multi_EN,fig:EF_IR_multi_DE,fig:EF_IR_multi_HI,fig:EF_IR_multi_ZH} and OLS estimates confirming these results are detailed in \Cref{tab:ols-coef-multi-en-jina-exp2,tab:ols-coef-multi-zh-jina-exp2,tab:ols-coef-multi-de-jina-exp2,tab:ols-coef-multi-hi-jina-exp2}.

% \footnote{Baseline segment-retentions of $n{=}3$ experiment instances for \textit{([it,\cEN{en},\cEN{...},\cEN{en}], [it,\cDE{de},\cDE{...},\cDE{de}])} are \textit{(\cEN{0.89}\sth, \cDE{0.92}\sth)} at position~1 ($\olsBetai{0}^{(\tau)}$); differences to baseline position~1 at position~2 ($\olsBetai{2}^{(\tau)}$) are \textit{(\cEN{-0.21}\sth, \cDEbold{-0.31}\sth)}. Detailed OLS estimates are shown in \Cref{tab:ols-coef-multi-en-jina-exp2,tab:ols-coef-multi-zh-jina-exp2,tab:ols-coef-multi-de-jina-exp2,tab:ols-coef-multi-hi-jina-exp2} in the Appendix.}

\begin{figure*}[t]
\centering
\includegraphics[width=0.969\textwidth]{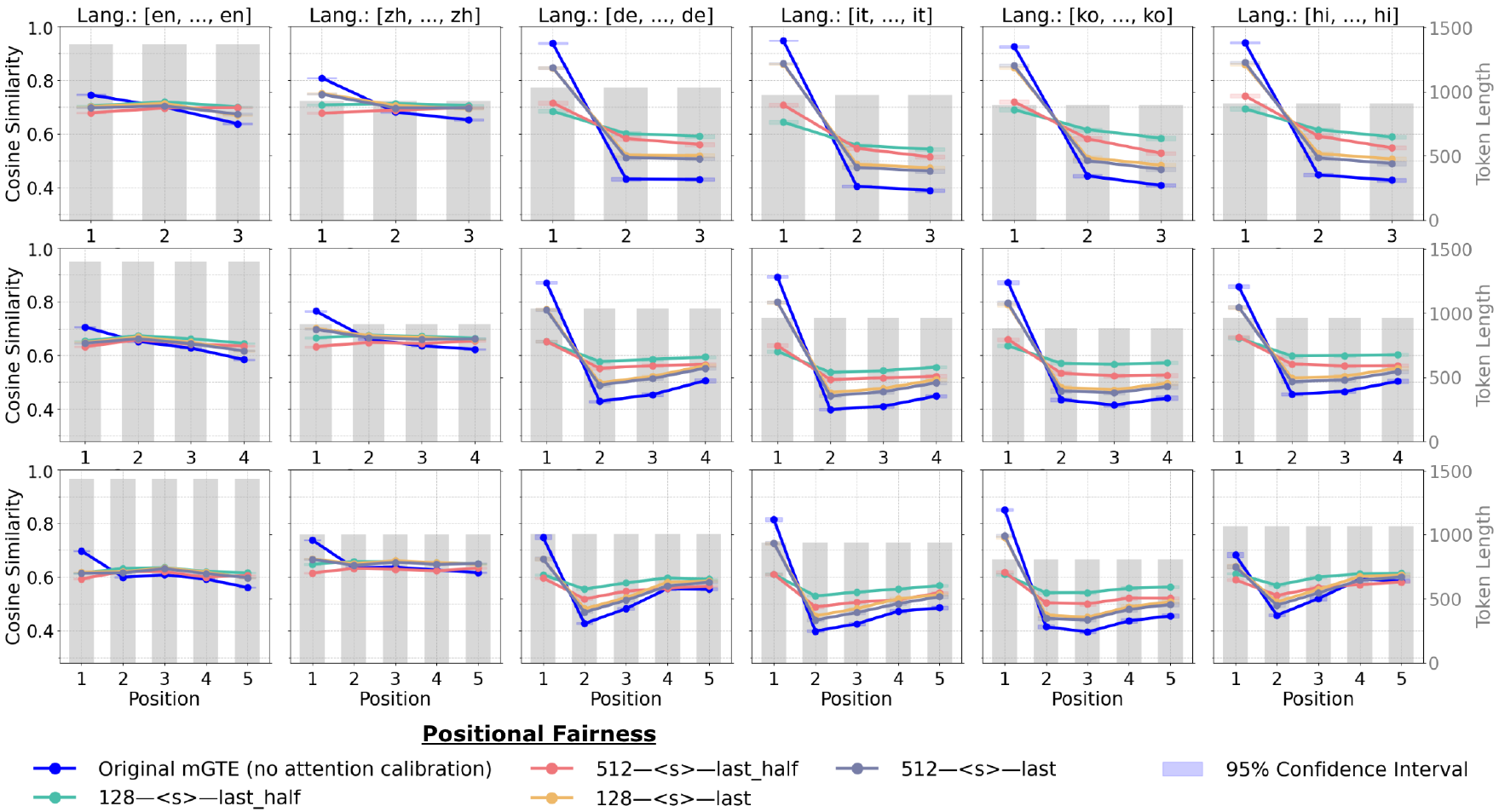}
  \caption{Monolingual document experiments using attention calibration on mGTE. Notation: \textit{128---$<$s$>$---last\_half}: Calibrated embeddings with hyperparameters $\basketSize{=}128$, $\layersCal{=}\{7,\ldots,12\}$}%  \textit{128---$<$s$>$---last}: $\basketSize{=}128$, $\layersCal{=}12$; \textit{512---$<$s$>$---last\_half}: $\basketSize{=}512$, $\layersCal{=}\{7,\ldots,12\}$; \textit{512---$<$s$>$---last}: $\basketSize{=}512$, $\layersCal{=}12$.}
  \label{fig:EF_mono_calibrated}
\end{figure*}

\subsubsection*{Analysis Overview}

As to positional fairness, independent of the embedding model, source languages, or document lengths, we observe a systematic \textbf{first-position bias}: the first segment is represented most strongly in the global document embedding. In mixed-language documents, position effects persist but are counteracted by \textbf{language preferences}: segments in English (and, for mGTE, Chinese) are represented well in the global document embedding, regardless of their position in the document.

Similarly, information retention is highest at the \textbf{first-positioned segment}. The magnitude of this effect \textbf{depends on language}. In monolingual documents, English shows smaller positional effects than German, Italian, Korean, and Hindi. In mixed-language documents, the first-position bias persists, but \textbf{English} segments retain much of their original representation, regardless of their position.

% do we want to add another general remark there?

\subsection{Self-Attention Distribution}
To examine potential origins of the positional bias, we analyze the self-attention distribution of the mGTE embedding model. We hypothesize a \emph{front-loaded} distribution over long inputs: early tokens receive more attention mass than later tokens, thereby exerting greater influence on intermediate representations and, ultimately, on the global document embedding. As the mGTE model uses $\poolingToken$-pooling, we examine how the $\poolingToken$-token query allocates attention mass over the sequence. We aggregate destinations (keys) into fixed-size, contiguous baskets of size 128 and show the $\poolingToken$- and $\eosToken$-token separately to avoid distorting the first and last basket by special tokens. If positional bias in global document representations is indeed driven by front-loaded self-attention distributions, we would expect L-shaped attention profiles.

\Cref{fig:attention_combined} shows that among content-bearing baskets, the first basket (tokens 2--129) receives the most attention from the $\poolingToken$-token. This aligns with the overrepresentation of the first-positioned segment in global document embeddings. Furthermore, slight mid/late-sequence increases in the attention profiles align with the U-shaped segment-representation profiles observed previously (cf. English in \Cref{fig:exp3_self_attn_mono_en_5seg_layerAVG} and Hindi in \Cref{fig:exp3_self_attn_mono_hi_5seg_layerAVG} in the Appendix). For some experiment instances, we observe partial agreement between self-attention profiles and the similarity of each segment's representation to the document representation; hence, we conclude that pooling-token self-attention can help explain positional bias patterns in document embeddings, but does not capture all nuances.

\section{Inference-Time Attention Calibration}

Motivated by the (partial) alignment between attention allocation and positional bias, we propose to re-weight attention scores during the forward pass. With this, we aim to counteract the front-loaded distribution, thereby allowing information from later tokens to be better represented in the global document embedding. This attention calibration operates entirely at inference time, i.e., it does not require any additional training.

\subsection*{Approach}

\begin{figure}[b!]
\centering
\includegraphics[width=0.953\linewidth, trim=5 2 2 2, clip]{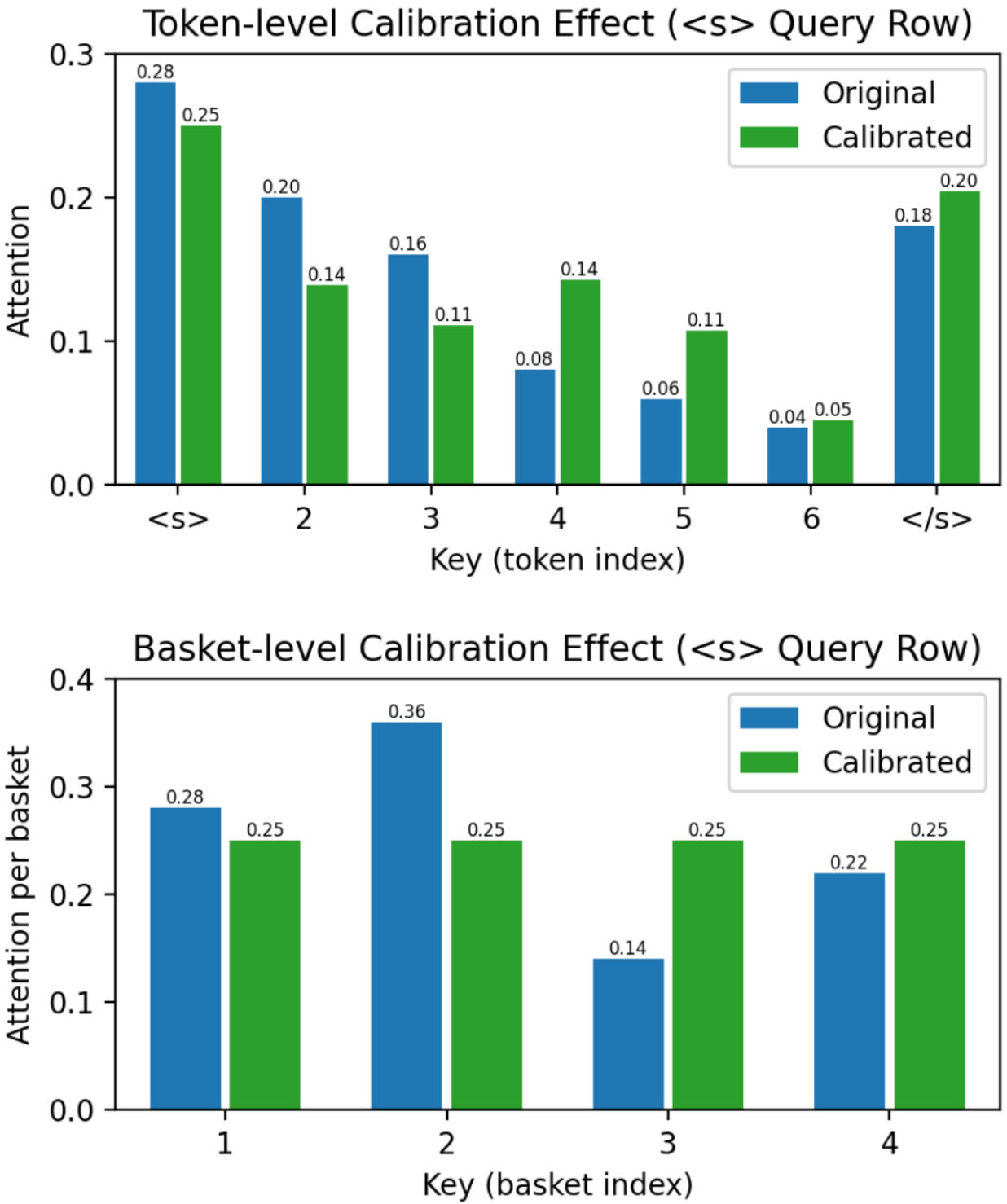} 
\caption{Conceptual illustration of the attention calibration. Calibration is applied to the $\poolingToken$-query token. The $\poolingToken$-key token is treated individually (basket index 1 contains only the $\poolingToken$-key token). Exemplarily, the remaining key sequence is grouped into baskets of size 2 (e.g., basket index 2 contains token indices 2 and 3).}
\label{fig:concept_attn_clb}
\end{figure}

\begin{figure*}[t]
\centering
\includegraphics[width=0.969\textwidth]{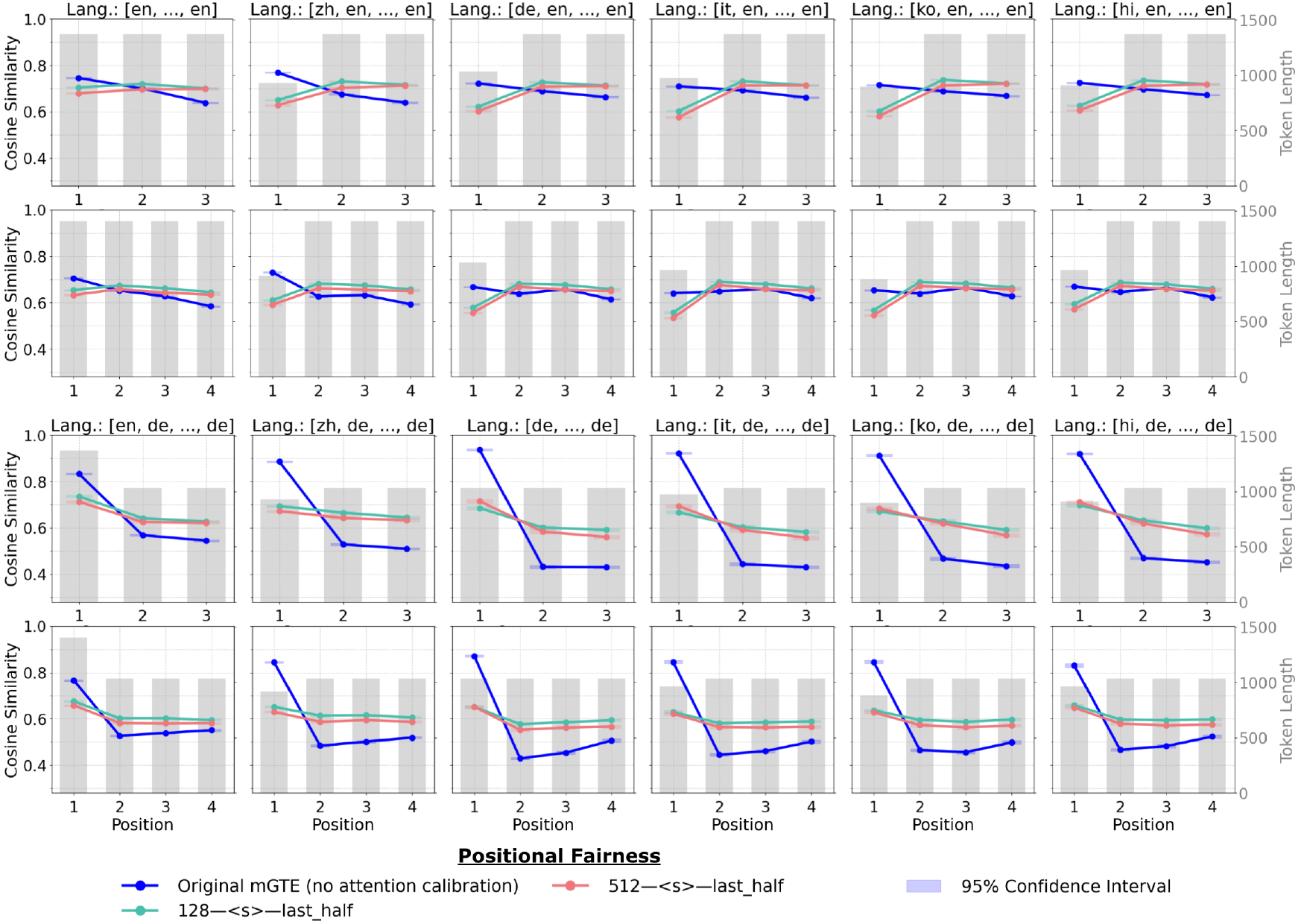}
\caption{Mixed-language document experiments using attention calibration on mGTE. Notation: \textit{128---$<$s$>$---last\_half}: Calibrated embeddings with hyperparameters $\basketSize{=}128$, $\layersCal{=}\{7,\ldots,12\}$}
% \caption{Using attention re-calibration exposes underlying language preferences of the model in the form of inverted L-shapes for $\langRemainder=en$ documents (top 2 rows). In contrast, $\langRemainder=de$ documents (bottom 2 rows) still exhibit slight L-shaped segment-representation profiles even after attention re-calibration. \textit{128---$<$s$>$---last\_half}: $\basketSize{=}128$, $\layersCal{=}\{7,\ldots,12\}$; \textit{512---$<$s$>$---last\_half}: $\basketSize{=}512$, $\layersCal{=}\{7,\ldots,12\}$.}
\label{fig:EF_multi_EN_DE_direct_comparison_calibrated}
\end{figure*}

Inspired by attention calibration for question answering in decoder models \citep{hsieh_found_2024}, we propose a different calibration technique for embedding models: inference-time attention mass equalization across positional baskets. Unlike \citet{hsieh_found_2024}, who require one inference per position, our method requires only two forward passes to obtain a calibrated embedding. We partition key positions into fixed-size, contiguous baskets and enforce a uniform attention mass for each basket, while preserving the relative distribution within each basket. This spreads the global attention budget of the $\poolingToken$-token over the full key sequence—later baskets receive as much total attention as earlier ones—while keeping fine-grained, within-basket patterns unchanged. \Cref{fig:concept_attn_clb} shows our approach on a conceptual level.

Our approach calibrates the $\poolingToken$-query row and takes two hyperparameters: (i) \textbf{basket size} ($\basketSize$): partitions the key sequence into $\basketSize$-sized baskets\footnote{We put the $\poolingToken$-pooling token in its own basket, yielding a total of $\lceil \frac{\numValidTokens-1}{\basketSize} \rceil+1$ baskets for an $\numValidTokens$-sized sequence.}; (ii) \textbf{calibrated layer set} ($\layersCal$): subset of transformer layers $\layersCal\subseteq\{1,\ldots,\numLayers\}$ where calibration is applied independently for each head.

\subsection*{Results: Monolingual Documents}

We apply attention calibration to mGTE and report results for the following hyperparameters: $\basketSize\in\{128,512\}$ and $\layersCal\in\{\{7,\ldots,12\},\{12\}\}$.

\Cref{fig:EF_mono_calibrated} shows a clear reduction of positional bias in all experiment instances when calibrated attention is used. This is achieved through two complementary adjustments: (i) the representation of the first-positioned segment is decreased relative to the uncalibrated baseline, counteracting its overrepresentation; (ii) the representations of later-positioned segments are increased, showing that their semantic content is better represented in the global document embedding after calibration. The OLS estimates confirm the substantial reduction of positional bias using calibrated embeddings, detailed in \Cref{tab:ols-coef-mono-alibaba-nlp-gte-multilingual-base-calibrated-comparison} in the Appendix.

%\footnote{Effect of attention re-calibration is shown as: \textit{original}$\rightarrow$\textit{re-calibrated}. Baseline segment-representations of $n{=}3$ experiment instances for \textit{(\cEN{en}, zh, \cDE{de}, it, ko, hi)} are  \textit{(\cEN{0.75}$\rightarrow$\cEN{0.70}, 0.81$\rightarrow$0.71, \cDE{0.94}$\rightarrow$\cDE{0.68}, 0.95$\rightarrow$0.64, 0.93$\rightarrow$0.69, 0.94$\rightarrow$0.69)} at position~1 ($\olsBetai{0}$); differences to baseline position~1 at position~2 ($\olsBetai{2}$) are \textit{(\cEN{-0.05}$\rightarrow$\cEN{0.02}, -0.13$\rightarrow$0.01, \cDEbold{-0.51}$\rightarrow$\cDEbold{-0.08}, -0.54$\rightarrow$-0.09, -0.48$\rightarrow$-0.07, -0.49$\rightarrow$-0.08)}. Detailed OLS estimates are shown in \Cref{tab:ols-coef-mono-alibaba-nlp-gte-multilingual-base-calibrated-comparison} in the Appendix.}

\subsection*{Results: Mixed-Language Documents}
We examine experiment instances with \(\langRemainder\in\{en,zh,de\}\), and use the following hyperparameters: $\basketSize\in\{128,512\}$ and $\layersCal{=}\{7,\ldots,12\}$.

\Cref{fig:EF_multi_EN_DE_direct_comparison_calibrated} shows experiment instances with $\langRemainder=en$ (top 2 rows) and corresponding instances with $\langRemainder=de$ (bottom 2 rows). Only for $\langRemainder=en$ (and, in weaker form, for $\langRemainder=zh$ in \Cref{fig:EF_multi_ZH_calibrated} in the Appendix), we observe an \textbf{inverted L-shape}. We interpret this as further evidence for the \textbf{English/Chinese language preferences} of the model: While in the uncalibrated model (\textcolor{Blue1}{blue}), the first-position bias partly offsets the language penalty of a non-English segment (e.g., German), the calibrated model reduces this positional uplift, leading to later English/Chinese segments being more strongly represented than the first-positioned segment. Detailed results of all language configurations are shown in \Cref{fig:EF_multi_EN_calibrated,fig:EF_multi_DE_calibrated,fig:EF_multi_ZH_calibrated} in the Appendix. The OLS estimates confirm these results in \Cref{tab:ols-coef-multi-en-mgte-calibrated-comparison,tab:ols-coef-multi-de-mgte-calibrated-comparison,tab:ols-coef-multi-zh-mgte-calibrated-comparison} in the Appendix.

Additionally, attention calibration reduces secondary positional effects, such as U-shaped tails. While for uncalibrated embeddings, we occasionally observe U-shaped segment-representation profiles for $\langRemainder=de$, attention calibration erases this curvature ($\olsBetai{2} \approx \olsBetai{3} \approx \olsBetai{4} \approx \olsBetai{5}$; cf. \Cref{tab:ols-coef-multi-de-mgte-calibrated-comparison} in the Appendix).

\subsection*{Evaluating the Loss of Semantics}

Semantics in the embedding space do not collapse. In fact, for later-positioned segments, the similarity between the document embedding and the standalone segment embedding is often equal to or higher under calibration than without, indicating that calibration preserves the original semantic content of the embeddings. To validate this finding, we conducted a control experiment for the two most aggressive calibration settings, calibrating only the document embeddings $\embedDoci{j}$ while keeping the standalone segment embeddings $\embedSegiDocjSTD{i}{j}$ unchanged. As shown in \Cref{fig:EF_mono_calibrated_DocOnly} in the Appendix, the unaltered embeddings exhibit near-identical or increased similarity with the calibrated document embeddings compared to the uncalibrated baseline. This indicates that calibration increases the meaningful information of later-positioned segments in the global document representation, as semantic degradation would have produced decreased similarities instead.

\section{Conclusions}
We investigate how position and language affect Information representation fairness when embedded texts are composed of multiple, thematically independent segments (e.g., newspapers). To this end, we provide (i) a diagnostic framework to quantify position–language interactions and positional fairness in long-document embeddings; (ii) clear empirical evidence that state-of-the-art encoder-based embedding models suffer from a first-position bias and model-dependent language preferences which both decrease fairness as to how long documents are embedded and how segments are contextualized; (iii) empirical evidence for front-loaded self-attention distributions that align with the observed first-position bias in global document embeddings; (iv) a training-free attention calibration method that substantially reduces positional biases by distributing representational capacity more evenly across a document. Together, they offer practitioners the tools to diagnose and reduce representational inequalities in page-level applications while offering a foundation for future work towards fairer multilingual long-context representations.

\section{Limitations}

%Our work has several limitations that point to directions for future research.

First, our notion of information representation fairness relies on cosine similarity between document embeddings and standalone segment embeddings as a proxy for how strongly a segment is reflected in the global representation. While this operationalization enables a tractable and model-agnostic evaluation, it does not directly measure intrinsic contribution of segments to the embedding. In particular, cosine similarity is sensitive to properties of the embedding space  and scaling, and thus to some extent may conflate representational alignment with true information contribution. Consequently, our fairness metric should be interpreted as a relative measure under a fixed probing representation, rather than a definitive characterization of segment-level influence.

Second, our analysis is restricted to \emph{encoder-based} long-context embedding models. While these models are widely used in embedding-based retrieval systems, decoder-based and hybrid architectures may exhibit different positional and language bias patterns. Extending the proposed evaluation framework to other embedding paradigms, including instruction-tuned or generative embedding models, remains an open direction.

Third, our evaluation relies on a multilingual comparable corpus constructed from Wikipedia articles. Although this design allows us to control semantic content across languages and systematically isolate positional effects, Wikipedia articles differ from real-world long documents such as reports, or legal texts in structure, discourse style, and noise. As a result, the absolute magnitude of the observed biases may differ in applied settings, even if the underlying mechanisms persist.

Fourth, we focus on representation-level effects and do not evaluate downstream task performance. Our notion of information representation fairness is defined geometrically—via cosine similarity and discoverability in embedding space—independent of any specific retrieval or ranking pipeline. While this abstraction enables task-agnostic diagnosis, future work should connect these findings to end-task outcomes in realistic retrieval systems. Early indications suggest the method is effective in downstream retrieval scenarios, but thorough evaluation remains for future work.

Fifth, our proposed attention calibration method is evaluated primarily on a single embedding model (mGTE) that uses pooling-token representations. While the method is conceptually model-agnostic and operates entirely at inference time, its effectiveness for other architectures and pooling strategies calls for further empirical validation by future work.

Finally, our study focuses on positional and language biases but does not consider other potential sources of representational inequality, such as topic frequency or domain mismatch. While prior work has examined how differences between naturally written text and LLM-rewritten text in various tones can affect embeddings \citep{writingstylematters2025}. Understanding how such biases interact with positional and language effects in long-document embeddings is an important avenue for future research.

\section*{Acknowledgments}
The authors received funding through the project \textit{Impresso – Media Monitoring of the Past II Beyond Borders: Connecting Historical Newspapers and Radio}. Impresso is a research project funded by the Swiss National Science Foundation (SNSF 213585) and the Luxembourg National Research Fund (17498891).

\bibliography{acl_latex}

\appendix
\section{Appendix}
\label{sec:appendix}

\begin{figure*}
  \centering
  \textbf{Sample of Wikipedia Comparable Corpus}\par\medskip
  \includegraphics[width=\textwidth]{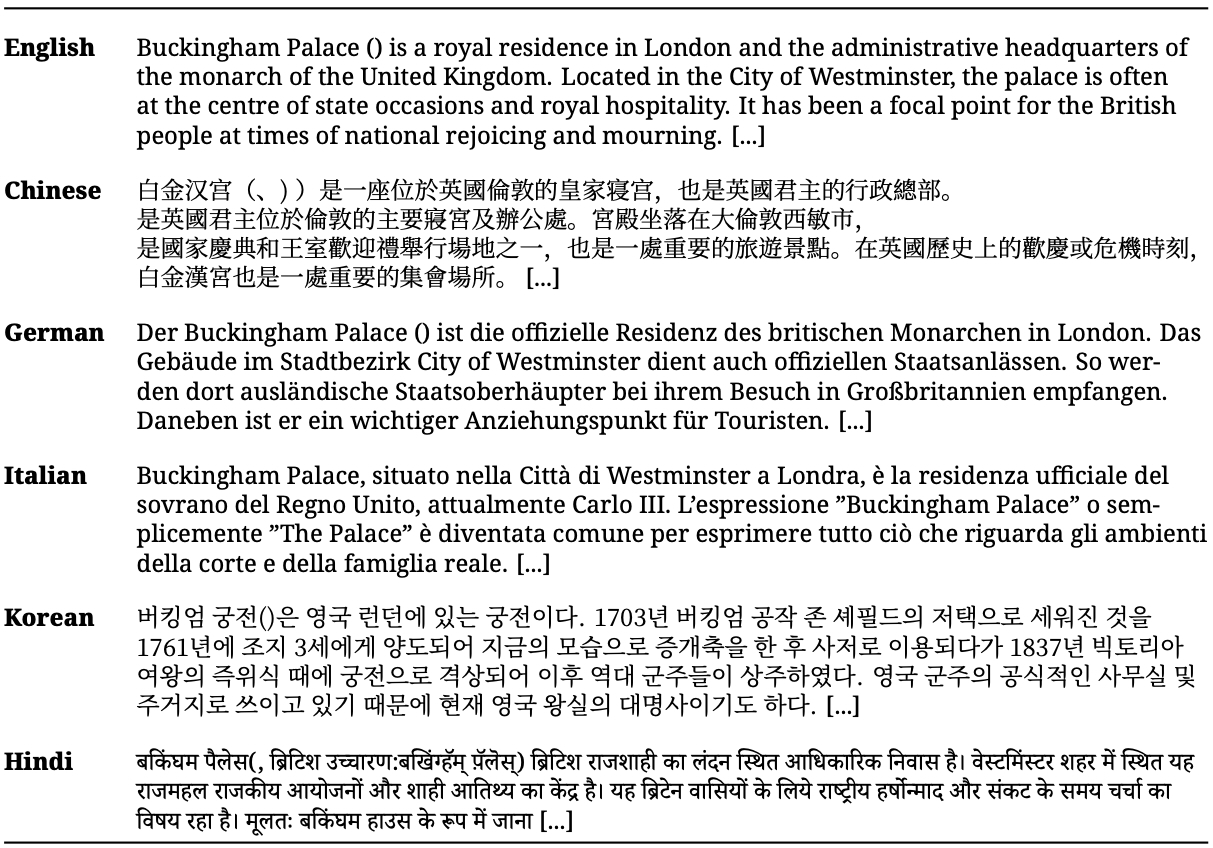}
  \caption[Sample of Wikipedia Comparable Corpus]{Sample of the multilingual Wikipedia comparable corpus across six languages. Exemplarily, the sample with ID 2059 (Wikipedia article about \textit{Buckingham Palace}) is shown. Every sample consists of the given article in all six languages. Articles are truncated for illustration purposes.}
  \label{fig:wiki-comparable-sample}
\end{figure*}

\begin{figure*}[tbp]
  \centering
  \includegraphics[width=\textwidth]{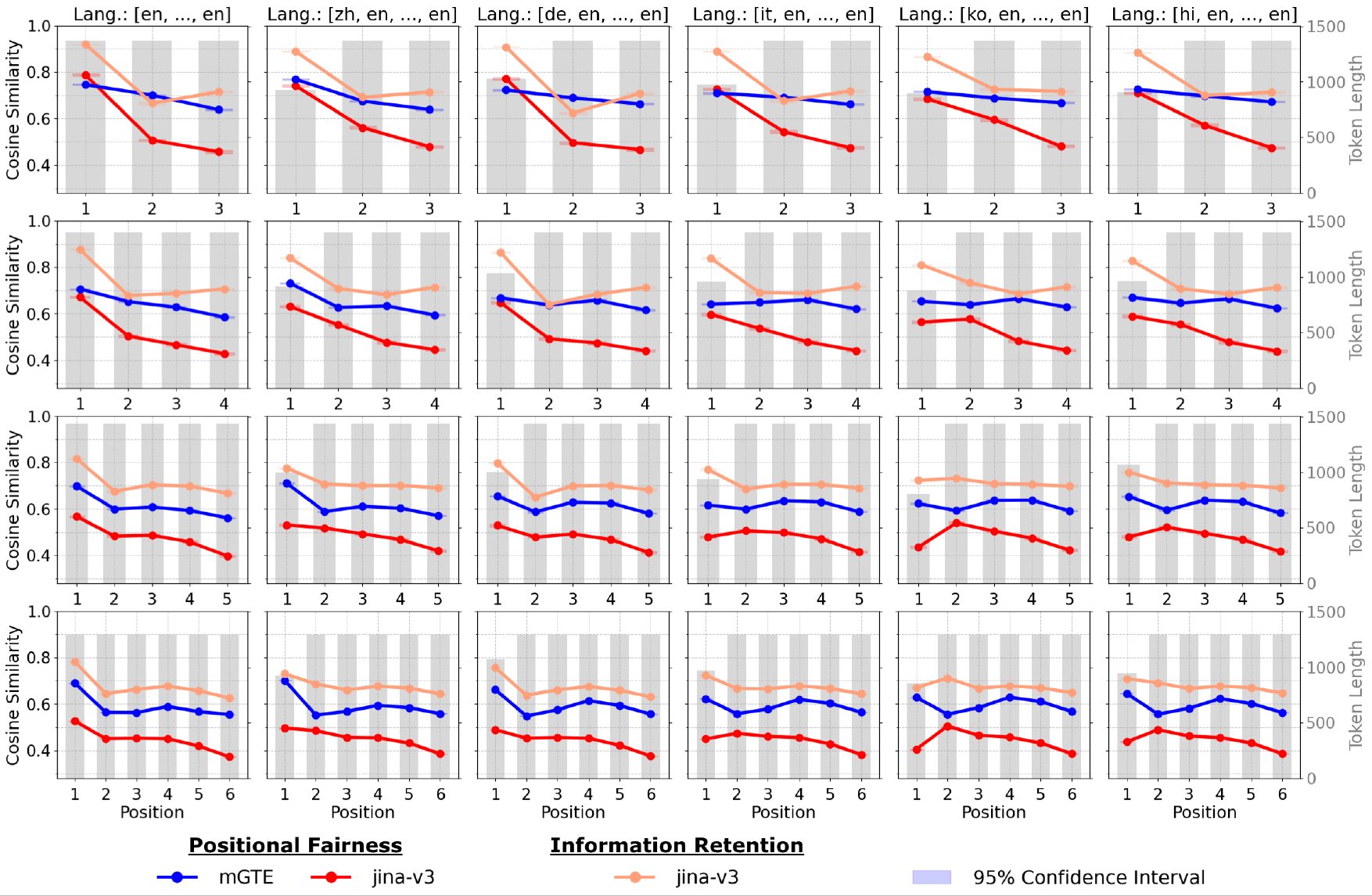}
  \caption{Subplots show different English ($\langRemainder=en$) mixed-language experiment instances $(n, \langConfig)$, where $n$ varies across rows, and $\langConfig$ varies across columns. Left y-axes show (i) average representation in the global document embedding (\textcolor{Blue1}{mGTE} and \textcolor{Red2}{jina-v3}), and (ii) average information retention (\textcolor{LightSalmon1}{jina-v3}) per segment position. Right y-axes show average token length per segment position (\textcolor{Snow4}{gray bars}).}
  \label{fig:EF_IR_multi_EN}
\end{figure*}

\begin{figure*}[tbp]
  \centering
  \includegraphics[width=\textwidth]{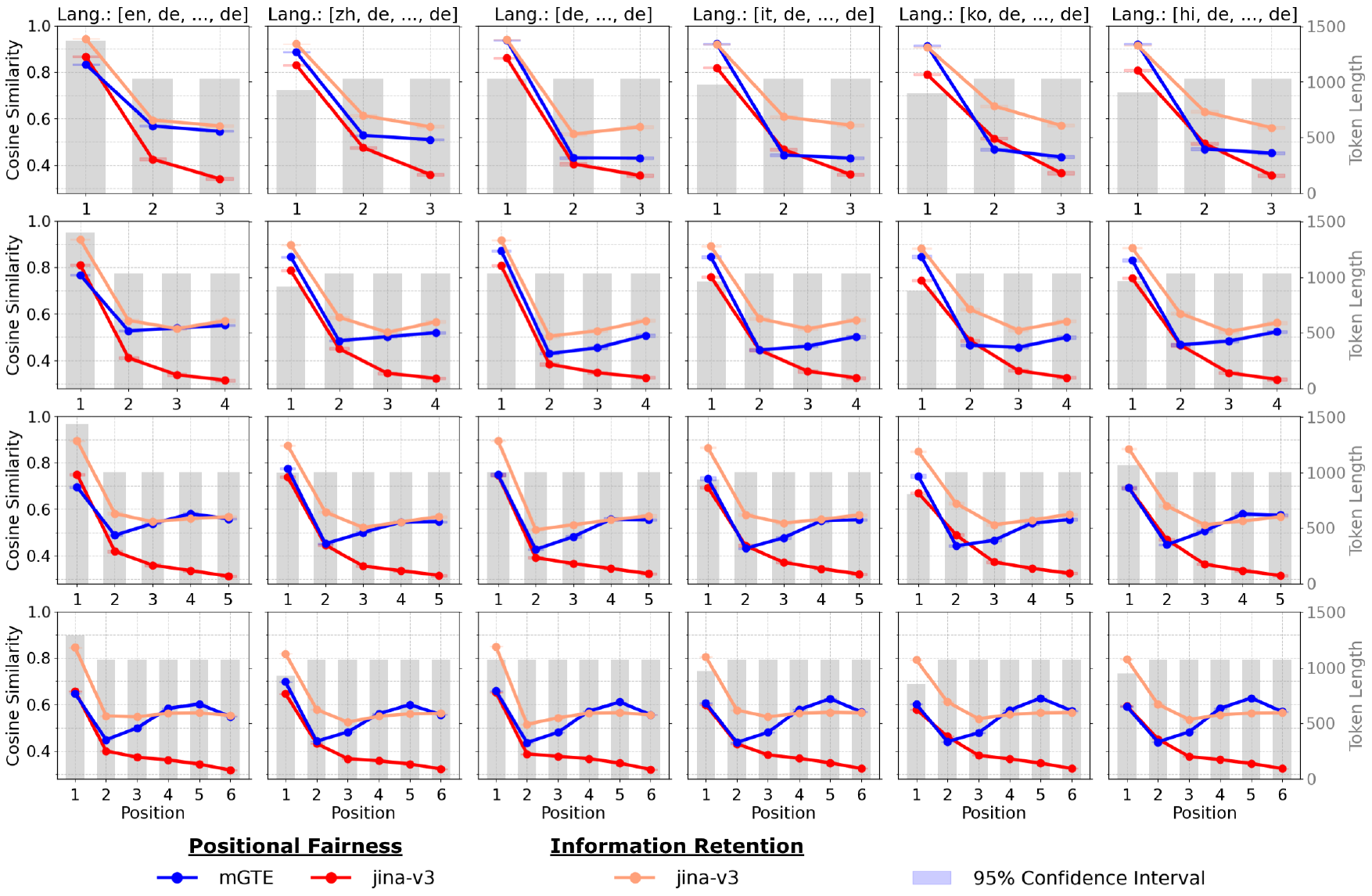}
  \caption{Subplots show different German ($\langRemainder=de$) mixed-language experiment instances $(n, \langConfig)$, where $n$ varies across rows, and $\langConfig$ varies across columns. Left y-axes show (i) average representation in the global document embedding (\textcolor{Blue1}{mGTE} and \textcolor{Red2}{jina-v3}), and (ii) average information retention (\textcolor{LightSalmon1}{jina-v3}) per segment position. Right y-axes show average token length per segment position (\textcolor{Snow4}{gray bars}).}
  \label{fig:EF_IR_multi_DE}
\end{figure*}

\begin{figure*}[tbp]
  \centering
  \includegraphics[width=\textwidth]{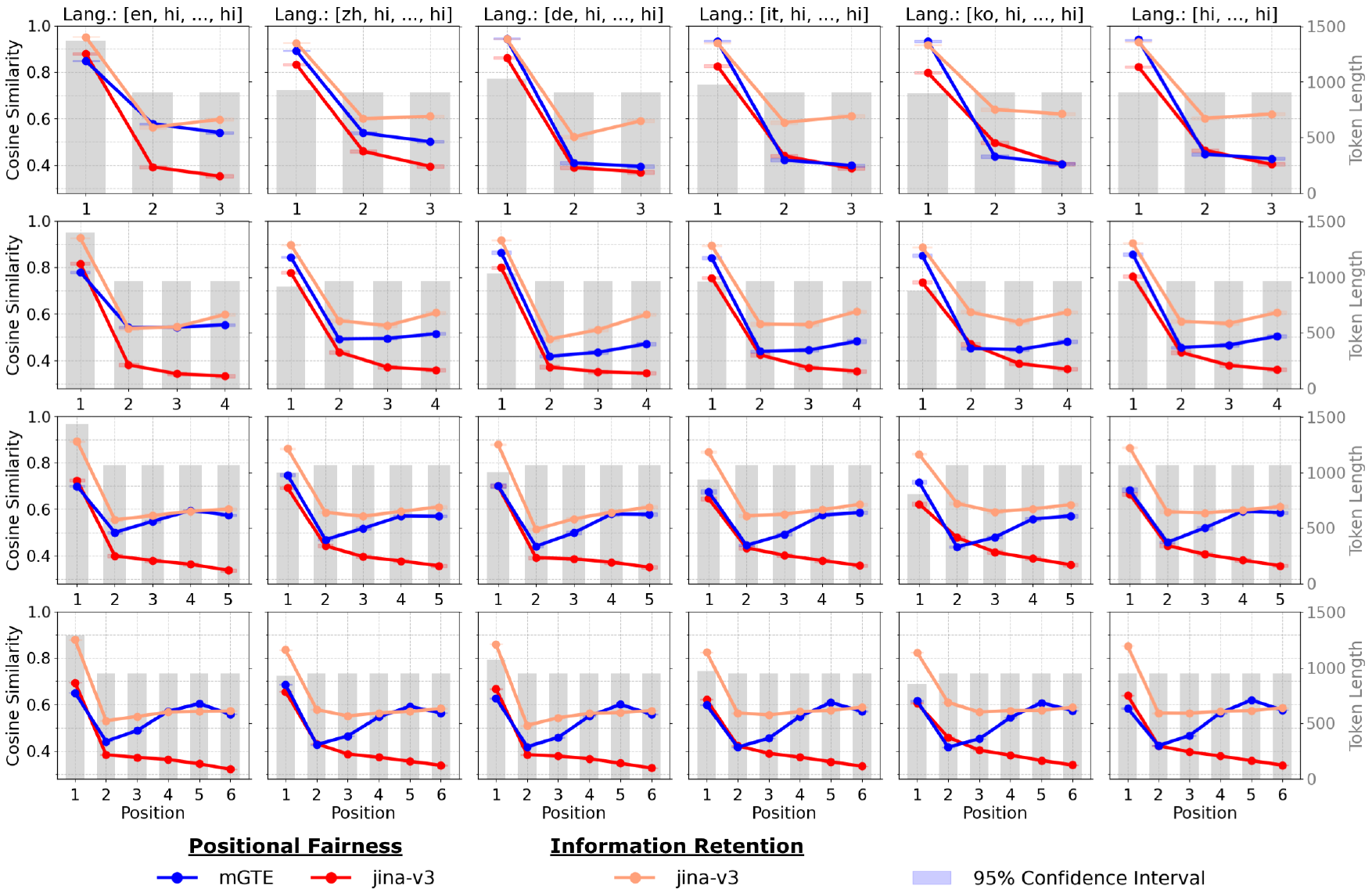}
  \caption{Subplots show different Hindi ($\langRemainder=hi$) mixed-language experiment instances $(n, \langConfig)$, where $n$ varies across rows, and $\langConfig$ varies across columns. Left y-axes show (i) average representation in the global document embedding (\textcolor{Blue1}{mGTE} and \textcolor{Red2}{jina-v3}), and (ii) average information retention (\textcolor{LightSalmon1}{jina-v3}) per segment position. Right y-axes show average token length per segment position (\textcolor{Snow4}{gray bars}).}
  \label{fig:EF_IR_multi_HI}
\end{figure*}

\begin{figure*}[tbp]
  \centering
  \includegraphics[width=\textwidth]{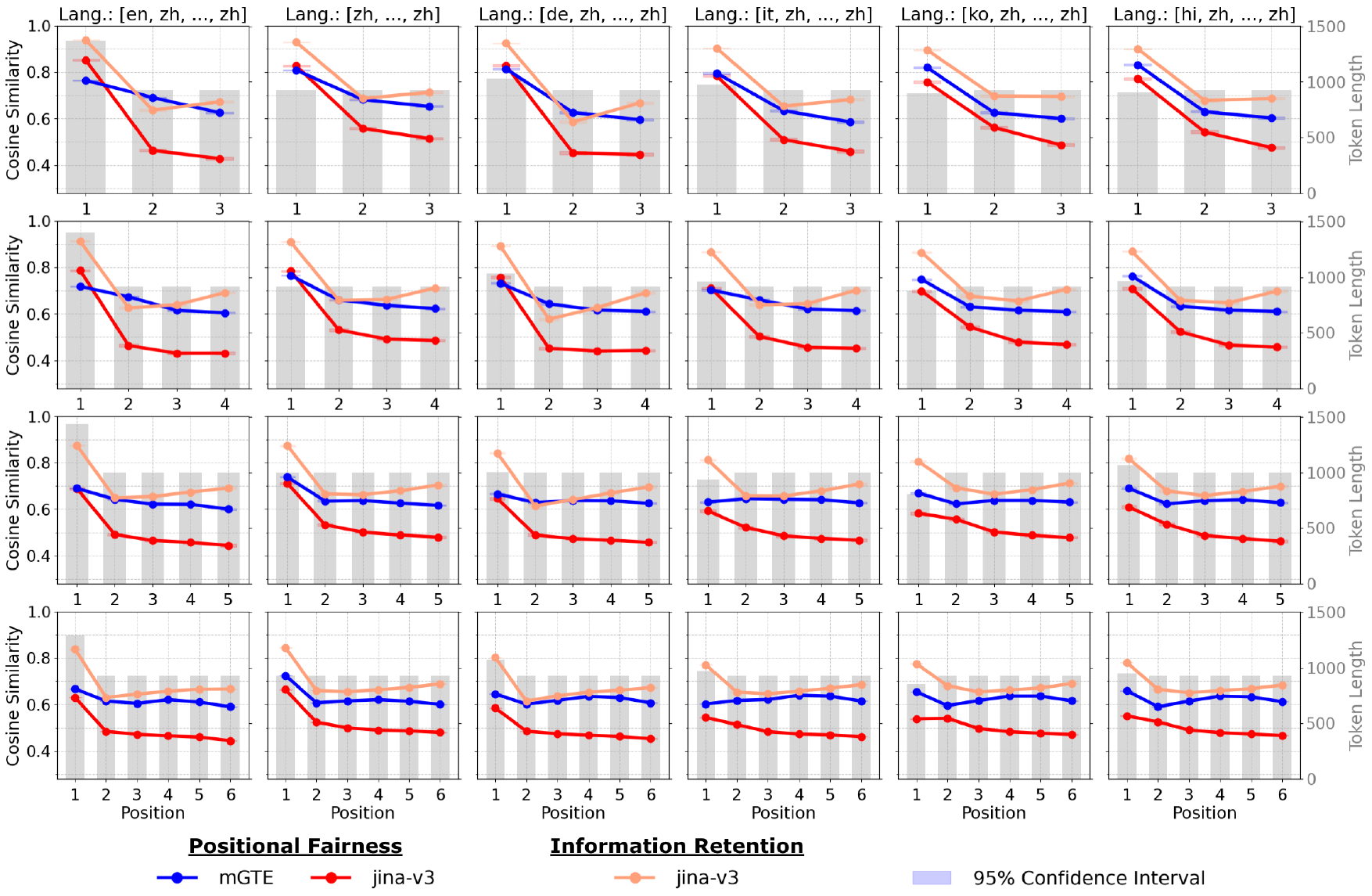}
  \caption{Subplots show different Chinese ($\langRemainder=zh$) mixed-language experiment instances $(n, \langConfig)$, where $n$ varies across rows, and $\langConfig$ varies across columns. Left y-axes show (i) average representation in the global document embedding (\textcolor{Blue1}{mGTE} and \textcolor{Red2}{jina-v3}), and (ii) average information retention (\textcolor{LightSalmon1}{jina-v3}) per segment position. Right y-axes show average token length per segment position (\textcolor{Snow4}{gray bars}).}
  \label{fig:EF_IR_multi_ZH}
\end{figure*}

\begin{table*}[]
  \centering
  \small
  \setlength{\tabcolsep}{6pt}
  \begin{tabular*}{\linewidth}{@{\extracolsep{\fill}} ll *{6}{c} @{}}
    \multicolumn{8}{c}{\shortstack{\textbf{OLS Coefficients Positional Fairness}\ \textbf{(monolingual, mGTE)}}} \\
    \addlinespace[2ex]
    \toprule
    \addlinespace[2ex]
    $n$& & en & zh & de & it & ko & hi \\
    \midrule
    \multirow{3}{*}{3} & $\olsBetai{0}$ $(\olsIntercept)$ & 0.75\sth & 0.81\sth & 0.94\sth & 0.95\sth & 0.93\sth & 0.94\sth \\
    & $\olsBetai{2}$ $(\olsPi{2})$ & \cEN{-0.05}\sth & -0.13\sth & \cDEbold{-0.51}\sth & -0.54\sth & -0.48\sth & -0.49\sth \\
    & $\olsBetai{3}$ $(\olsPi{3})$ & \cEN{-0.11}\sth & -0.16\sth & \cDEbold{-0.51}\sth & -0.56\sth & -0.52\sth & -0.51\sth \\
    \addlinespace[2ex]
    \multirow{4}{*}{4} & $\olsBetai{0}$ $(\olsIntercept)$ & 0.71\sth & 0.77\sth & 0.87\sth & 0.89\sth & 0.87\sth & 0.86\sth \\
    & $\olsBetai{2}$ $(\olsPi{2})$ & -0.05\sth & -0.11\sth & -0.44\sth & -0.49\sth & -0.44\sth & -0.40\sth \\
    & $\olsBetai{3}$ $(\olsPi{3})$ & -0.08\sth & -0.13\sth & -0.42\sth & -0.48\sth & -0.46\sth & -0.39\sth \\
    & $\olsBetai{4}$ $(\olsPi{4})$ & -0.12\sth & -0.14\sth & -0.36\sth & -0.44\sth & -0.43\sth & -0.35\sth \\
    \addlinespace[2ex]
    \multirow{5}{*}{5} & $\olsBetai{0}$ $(\olsIntercept)$ & 0.70\sth & 0.74\sth & 0.75\sth & 0.82\sth & 0.85\sth & 0.68\sth \\
    & $\olsBetai{2}$ $(\olsPi{2})$ & -0.10\sth & -0.10\sth & -0.32\sth & -0.42\sth & -0.44\sth & -0.22\sth \\
    & $\olsBetai{3}$ $(\olsPi{3})$ & -0.09\sth & -0.10\sth & -0.27\sth & -0.39\sth & -0.46\sth & -0.16\son \\
    & $\olsBetai{4}$ $(\olsPi{4})$ & -0.10\sth & -0.11\sth & -0.19\son & -0.34\sth & -0.41\sth & -0.09\snn \\
    & $\olsBetai{5}$ $(\olsPi{5})$ & -0.14\sth & -0.12\sth & -0.19\stw & -0.33\sth & -0.40\sth & -0.10\snn \\
    \addlinespace[2ex]
    \multirow{6}{*}{6} & $\olsBetai{0}$ $(\olsIntercept)$ & 0.69\sth & 0.72\sth & 0.66\sth & 0.66\sth & 0.71\sth & 0.58\sth \\
    & $\olsBetai{2}$ $(\olsPi{2})$ & \cEN{-0.13}\sth & -0.12\sth & \cDEbold{-0.22}\sth & -0.24\sth & -0.28\sth & -0.16\sth \\
    & $\olsBetai{3}$ $(\olsPi{3})$ & \cEN{-0.13}\sth & -0.11\sth & \cDEbold{-0.18}\sth & -0.21\sth & -0.26\sth & -0.12\son \\
    & $\olsBetai{4}$ $(\olsPi{4})$ & \cENbold{-0.10}\sth & -0.10\sth & \cDE{-0.09}\stw & -0.11\son & -0.21\sth & -0.02\snn \\
    & $\olsBetai{5}$ $(\olsPi{5})$ & \cEN{-0.12}\sth & -0.11\sth & -0.05\snn & -0.07\snn & -0.14\son & 0.03\snn \\
    & $\olsBetai{6}$ $(\olsPi{6})$ & \cENbold{-0.14}\sth & -0.12\sth & \cDE{-0.10}\sth & -0.12\stw & -0.13\stw & -0.01\snn \\
    \bottomrule
  \end{tabular*}
  \caption{Estimated OLS coefficients of the positional fairness analysis (similarity between document embedding and standalone segment embeddings) in the monolingual document setting, using mGTE embeddings. $n$ denotes the number of segments per document. OLS regression uses positions as categorical variables; $\olsBetai{0}$ captures the baseline (similarity between document embedding and standalone embedding at position~1), $\olsBetai{p\ge2}$ captures the difference between position $p$'s similarity (similarity between document embedding and standalone embedding at position~$p$) and the baseline similarity. Statistical significance: all values $\mathfrak{p}<0.001$; unless indicated otherwise: \stw $\mathfrak{p}<0.01$, \son $\mathfrak{p}<0.05$, \snn not significant.}
  \label{tab:ols-coef-mono-mgte}
\end{table*}

\begin{table*}[]
  \centering
  \small
  \setlength{\tabcolsep}{6pt}
  \begin{tabular*}{\linewidth}{@{\extracolsep{\fill}} ll *{6}{c} @{}}
    \multicolumn{8}{c}{\shortstack{\textbf{OLS Coefficients Positional Fairness}\ \textbf{(monolingual, jina-v3)}}} \\
    \addlinespace[2ex]
    \toprule
    \addlinespace[2ex]
    $n$& & en & zh & de & it & ko & hi \\
    \midrule
    \multirow{3}{*}{3} & $\olsBetai{0}$ $(\olsIntercept)$ & 0.79\sth & 0.83\sth & 0.86\sth & 0.82\sth & 0.79\sth & 0.82\sth \\
    & $\olsBetai{2}$ $(\olsPi{2})$ & -0.28\sth & -0.27\sth & -0.46\sth & -0.37\sth & -0.30\sth & -0.36\sth \\
    & $\olsBetai{3}$ $(\olsPi{3})$ & -0.33\sth & -0.31\sth & -0.50\sth & -0.43\sth & -0.35\sth & -0.42\sth \\
    \addlinespace[2ex]
    \multirow{4}{*}{4} & $\olsBetai{0}$ $(\olsIntercept)$ & 0.67\sth & 0.78\sth & 0.81\sth & 0.76\sth & 0.74\sth & 0.76\sth \\
    & $\olsBetai{2}$ $(\olsPi{2})$ & -0.17\sth & -0.25\sth & -0.42\sth & -0.35\sth & -0.30\sth & -0.33\sth \\
    & $\olsBetai{3}$ $(\olsPi{3})$ & -0.20\sth & -0.29\sth & -0.46\sth & -0.40\sth & -0.35\sth & -0.38\sth \\
    & $\olsBetai{4}$ $(\olsPi{4})$ & -0.24\sth & -0.30\sth & -0.48\sth & -0.42\sth & -0.36\sth & -0.40\sth \\
    \addlinespace[2ex]
    \multirow{5}{*}{5} & $\olsBetai{0}$ $(\olsIntercept)$ & 0.57\sth & 0.71\sth & 0.75\sth & 0.70\sth & 0.70\sth & 0.67\sth \\
    & $\olsBetai{2}$ $(\olsPi{2})$ & -0.08\sth & -0.18\sth & -0.36\sth & -0.29\sth & -0.27\sth & -0.22\sth \\
    & $\olsBetai{3}$ $(\olsPi{3})$ & -0.08\sth & -0.21\sth & -0.38\sth & -0.33\sth & -0.33\sth & -0.26\sth \\
    & $\olsBetai{4}$ $(\olsPi{4})$ & -0.11\sth & -0.22\sth & -0.40\sth & -0.36\sth & -0.34\sth & -0.28\sth \\
    & $\olsBetai{5}$ $(\olsPi{5})$ & -0.17\sth & -0.23\sth & -0.42\sth & -0.37\sth & -0.34\sth & -0.31\sth \\
    \addlinespace[2ex]
    \multirow{6}{*}{6} & $\olsBetai{0}$ $(\olsIntercept)$ & 0.53\sth & 0.66\sth & 0.65\sth & 0.62\sth & 0.63\sth & 0.64\sth \\
    & $\olsBetai{2}$ $(\olsPi{2})$ & -0.08\sth & -0.14\sth & -0.26\sth & -0.20\sth & -0.20\sth & -0.22\sth \\
    & $\olsBetai{3}$ $(\olsPi{3})$ & -0.07\sth & -0.16\sth & -0.27\sth & -0.23\sth & -0.23\sth & -0.24\sth \\
    & $\olsBetai{4}$ $(\olsPi{4})$ & -0.08\sth & -0.17\sth & -0.28\sth & -0.25\sth & -0.25\sth & -0.26\sth \\
    & $\olsBetai{5}$ $(\olsPi{5})$ & -0.11\sth & -0.18\sth & -0.30\sth & -0.27\sth & -0.26\sth & -0.28\sth \\
    & $\olsBetai{6}$ $(\olsPi{6})$ & -0.15\sth & -0.18\sth & -0.33\sth & -0.29\sth & -0.27\sth & -0.30\sth \\
    \bottomrule
  \end{tabular*}
  \caption{Estimated OLS coefficients of the positional fairness analysis (similarity between document embedding and standalone segment embeddings) in the monolingual document setting, using jina-v3 embeddings. $n$ denotes the number of segments per document. OLS regression uses positions as categorical variables; $\olsBetai{0}$ captures the baseline (similarity between document embedding and standalone embedding at position~1), $\olsBetai{p\ge2}$ captures the difference between position $p$'s similarity (similarity between document embedding and standalone embedding at position~$p$) and the baseline similarity. Statistical significance: all values $\mathfrak{p}<0.001$.}
  \label{tab:ols-coef-mono-jina}
\end{table*}

\begin{table*}[tb]
  \centering
  \small
  \setlength{\tabcolsep}{6pt}
  \begin{tabular*}{\linewidth}{@{\extracolsep{\fill}} ll *{5}{c} @{}}
    \multicolumn{7}{c}{\shortstack{\textbf{OLS Coefficients Positional Fairness}\\ \textbf{(mixed-language English, mGTE)}}} \\
    \addlinespace[2ex]
    \toprule
    \addlinespace[2ex]
    $n$& & \shortstack{[zh, en,\\ \ldots, en]} & \shortstack{[de, en,\\ \ldots, en]} & \shortstack{[it, en,\\ \ldots, en]} & \shortstack{[ko, en,\\ \ldots, en]} & \shortstack{[hi, en,\\ \ldots, en]} \\
    \midrule
    \multirow{3}{*}{3} & $\olsBetai{0}$ $(\olsIntercept)$ & 0.77\sth & 0.72\sth & 0.71\sth & 0.72\sth & 0.72\sth \\
    & $\olsBetai{2}$ $(\olsPi{2})$ & -0.09\sth & -0.03\sth & -0.02\sth & -0.03\sth & -0.03\sth \\
    & $\olsBetai{3}$ $(\olsPi{3})$ & -0.13\sth & -0.06\sth & -0.05\sth & -0.05\sth & -0.05\sth \\
    \addlinespace[2ex]
    \multirow{4}{*}{4} & $\olsBetai{0}$ $(\olsIntercept)$ & 0.73\sth & 0.67\sth & 0.64\sth & 0.65\sth & 0.67\sth \\
    & $\olsBetai{2}$ $(\olsPi{2})$ & -0.10\sth & -0.03\sth & 0.01\snn & -0.01\stw & -0.02\sth \\
    & $\olsBetai{3}$ $(\olsPi{3})$ & -0.10\sth & -0.01\son & 0.02\sth & 0.01\son & -0.01\snn \\
    & $\olsBetai{4}$ $(\olsPi{4})$ & -0.14\sth & -0.05\sth & -0.02\sth & -0.03\sth & -0.05\sth \\
    \addlinespace[2ex]
    \multirow{5}{*}{5} & $\olsBetai{0}$ $(\olsIntercept)$ & 0.71\sth & 0.65\sth & 0.62\sth & 0.62\sth & 0.65\sth \\
    & $\olsBetai{2}$ $(\olsPi{2})$ & -0.12\sth & -0.07\sth & -0.02\snn & -0.03\stw & -0.06\sth \\
    & $\olsBetai{3}$ $(\olsPi{3})$ & -0.10\sth & -0.03\sth & 0.02\snn & 0.01\snn & -0.02\snn \\
    & $\olsBetai{4}$ $(\olsPi{4})$ & -0.11\sth & -0.03\sth & 0.01\snn & 0.01\snn & -0.02\son\\
    & $\olsBetai{5}$ $(\olsPi{5})$ & -0.14\sth & -0.07\sth & -0.03\stw & -0.03\sth & -0.07\sth \\
    \addlinespace[2ex]
    \multirow{6}{*}{6} & $\olsBetai{0}$ $(\olsIntercept)$ & 0.70\sth & 0.66\sth & 0.62\sth & 0.63\sth & 0.64\sth \\
    & $\olsBetai{2}$ $(\olsPi{2})$ & -0.15\sth & -0.11\sth & -0.06\sth & -0.07\sth & -0.09\sth \\
    & $\olsBetai{3}$ $(\olsPi{3})$ & -0.13\sth & -0.09\sth & -0.04\sth & -0.04\sth & -0.06\sth \\
    & $\olsBetai{4}$ $(\olsPi{4})$ & -0.11\sth & -0.05\sth & -0.00\snn & 0.00\snn & -0.02\sth \\
    & $\olsBetai{5}$ $(\olsPi{5})$ & -0.12\sth & -0.07\sth & -0.02\stw & -0.02\stw & -0.04\sth \\
    & $\olsBetai{6}$ $(\olsPi{6})$ & -0.14\sth & -0.11\sth & -0.06\sth & -0.06\sth & -0.08\sth \\
    \bottomrule
  \end{tabular*}
  \caption[Exp. 1: OLS Results. Mixed-Language (En), mGTE.]{Estimated OLS coefficients of the positional fairness analysis (similarity between document embedding and standalone segment embeddings) in the mixed-language document setting (\textbf{English}; $\langRemainder=en$), using mGTE embeddings. $n$ denotes the number of segments per document. OLS regression uses positions as categorical variables; $\olsBetai{0}$ captures the baseline (similarity between document embedding and standalone embedding at position~1), $\olsBetai{p\ge2}$ captures the difference between position $p$'s similarity (similarity between document embedding and standalone embedding at position~$p$) and the baseline similarity. Statistical significance: all values $\mathfrak{p}<0.001$; unless indicated otherwise: \stw $\mathfrak{p}<0.01$, \son $\mathfrak{p}<0.05$, \snn not significant.}
  \label{tab:ols-coef-multi-en-mgte}
\end{table*}

\begin{table*}[tb]
  \centering
  \small
  \setlength{\tabcolsep}{6pt}
  \begin{tabular*}{\linewidth}{@{\extracolsep{\fill}} ll *{5}{c} @{}}
    \multicolumn{7}{c}{\shortstack{\textbf{OLS Coefficients Positional Fairness}\\ \textbf{(mixed-language German, mGTE)}}} \\
    \addlinespace[2ex]
    \toprule
    \addlinespace[2ex]
    $n$& & \shortstack{[en, de,\\ \ldots, de]} & \shortstack{[zh, de,\\ \ldots, de]} & \shortstack{[it, de,\\ \ldots, de]} & \shortstack{[ko, de,\\ \ldots, de]} & \shortstack{[hi, de,\\ \ldots, de]} \\
    \midrule
    \multirow{3}{*}{3} & $\olsBetai{0}$ $(\olsIntercept)$ & 0.83\sth & 0.89\sth & 0.92\sth & 0.91\sth & 0.92\sth \\
    & $\olsBetai{2}$ $(\olsPi{2})$ & -0.26\sth & -0.36\sth & -0.48\sth & -0.45\sth & -0.45\sth \\
    & $\olsBetai{3}$ $(\olsPi{3})$ & -0.29\sth & -0.38\sth & -0.49\sth & -0.48\sth & -0.47\sth \\
    \addlinespace[2ex]
    \multirow{4}{*}{4} & $\olsBetai{0}$ $(\olsIntercept)$ & 0.77\sth & 0.84\sth & 0.84\sth & 0.85\sth & 0.83\sth \\
    & $\olsBetai{2}$ $(\olsPi{2})$ & -0.24\sth & -0.36\sth & -0.40\sth & -0.38\sth & -0.36\sth \\
    & $\olsBetai{3}$ $(\olsPi{3})$ & -0.23\sth & -0.34\sth & -0.38\sth & -0.39\sth & -0.35\sth \\
    & $\olsBetai{4}$ $(\olsPi{4})$ & -0.22\sth & -0.33\sth & -0.34\sth & -0.35\sth & -0.31\sth \\
    \addlinespace[2ex]
    \multirow{5}{*}{5} & $\olsBetai{0}$ $(\olsIntercept)$ & 0.69\sth & 0.78\sth & 0.73\sth & 0.74\sth & 0.69\sth \\
    & $\olsBetai{2}$ $(\olsPi{2})$ & -0.21\sth & -0.32\sth & -0.30\sth & -0.30\sth & -0.24\sth \\
    & $\olsBetai{3}$ $(\olsPi{3})$ & -0.16\sth & -0.28\sth & -0.25\sth & -0.28\sth & -0.19\sth \\
    & $\olsBetai{4}$ $(\olsPi{4})$ & -0.11\sth & -0.23\sth & -0.18\sth & -0.20\sth & -0.11\stw \\
    & $\olsBetai{5}$ $(\olsPi{5})$ & -0.14\sth & -0.23\sth & -0.18\sth & -0.19\sth & -0.12\sth \\
    \addlinespace[2ex]
    \multirow{6}{*}{6} & $\olsBetai{0}$ $(\olsIntercept)$ & 0.65\sth & 0.70\sth & 0.61\sth & 0.60\sth & 0.59\sth \\
    & $\olsBetai{2}$ $(\olsPi{2})$ & -0.20\sth & -0.26\sth & -0.17\sth & -0.16\sth & -0.15\sth \\
    & $\olsBetai{3}$ $(\olsPi{3})$ & -0.15\sth & -0.21\sth & -0.13\sth & -0.12\sth & -0.11\sth \\
    & $\olsBetai{4}$ $(\olsPi{4})$ & -0.06\sth & -0.14\sth & -0.03\snn & -0.03\snn & -0.01\snn \\
    & $\olsBetai{5}$ $(\olsPi{5})$ & -0.04\sth & -0.10\sth & 0.02\snn & 0.03\snn & 0.04\son\\
    & $\olsBetai{6}$ $(\olsPi{6})$ & -0.10\sth & -0.14\sth & -0.04\stw & -0.03\son & -0.02\snn \\
    \bottomrule
  \end{tabular*}
  \caption[Exp. 1: OLS Results. Mixed-Language (De), mGTE.]{Estimated OLS coefficients of the positional fairness analysis (similarity between document embedding and standalone segment embeddings) in the mixed-language document setting (\textbf{German}; $\langRemainder=de$), using mGTE embeddings. $n$ denotes the number of segments per document. OLS regression uses positions as categorical variables; $\olsBetai{0}$ captures the baseline (similarity between document embedding and standalone embedding at position~1), $\olsBetai{p\ge2}$ captures the difference between position $p$'s similarity (similarity between document embedding and standalone embedding at position~$p$) and the baseline similarity. Statistical significance: all values $\mathfrak{p}<0.001$; unless indicated otherwise: \stw $\mathfrak{p}<0.01$, \son $\mathfrak{p}<0.05$, \snn not significant.}
  \label{tab:ols-coef-multi-de-mgte}
\end{table*}

\begin{table*}[tb]
  \centering
  \small
  \setlength{\tabcolsep}{6pt}
  \begin{tabular*}{\linewidth}{@{\extracolsep{\fill}} ll *{5}{c} @{}}
    \multicolumn{7}{c}{\shortstack{\textbf{OLS Coefficients Positional Fairness}\\ \textbf{(mixed-language Hindi, mGTE)}}} \\
    \addlinespace[2ex]
    \toprule
    \addlinespace[2ex]
    $n$& & \shortstack{[en, hi,\\ \ldots, hi]} & \shortstack{[zh, hi,\\ \ldots, hi]} & \shortstack{[de, hi,\\ \ldots, hi]} & \shortstack{[it, hi,\\ \ldots, hi]} & \shortstack{[ko, hi,\\ \ldots, hi]} \\
    \midrule
    \multirow{3}{*}{3} & $\olsBetai{0}$ $(\olsIntercept)$ & 0.85\sth & 0.89\sth & 0.94\sth & 0.93\sth & 0.93\sth \\
    & $\olsBetai{2}$ $(\olsPi{2})$ & -0.27\sth & -0.35\sth & -0.53\sth & -0.51\sth & -0.49\sth \\
    & $\olsBetai{3}$ $(\olsPi{3})$ & -0.31\sth & -0.39\sth & -0.55\sth & -0.53\sth & -0.53\sth \\
    \addlinespace[2ex]
    \multirow{4}{*}{4} & $\olsBetai{0}$ $(\olsIntercept)$ & 0.78\sth & 0.84\sth & 0.86\sth & 0.84\sth & 0.85\sth \\
    & $\olsBetai{2}$ $(\olsPi{2})$ & -0.24\sth & -0.35\sth & -0.45\sth & -0.40\sth & -0.40\sth \\
    & $\olsBetai{3}$ $(\olsPi{3})$ & -0.24\sth & -0.35\sth & -0.43\sth & -0.40\sth & -0.40\sth \\
    & $\olsBetai{4}$ $(\olsPi{4})$ & -0.23\sth & -0.33\sth & -0.39\sth & -0.36\sth & -0.37\sth \\
    \addlinespace[2ex]
    \multirow{5}{*}{5} & $\olsBetai{0}$ $(\olsIntercept)$ & 0.70\sth & 0.75\sth & 0.70\sth & 0.67\sth & 0.72\sth \\
    & $\olsBetai{2}$ $(\olsPi{2})$ & -0.20\sth & -0.28\sth & -0.26\sth & -0.23\sth & -0.28\sth \\
    & $\olsBetai{3}$ $(\olsPi{3})$ & -0.15\sth & -0.23\sth & -0.20\sth & -0.18\sth & -0.24\sth \\
    & $\olsBetai{4}$ $(\olsPi{4})$ & -0.10\sth & -0.18\sth & -0.12\stw & -0.10\son & -0.16\sth \\
    & $\olsBetai{5}$ $(\olsPi{5})$ & -0.12\sth & -0.18\sth & -0.12\sth & -0.09\son & -0.15\sth \\
    \addlinespace[2ex]
    \multirow{6}{*}{6} & $\olsBetai{0}$ $(\olsIntercept)$ & 0.65\sth & 0.69\sth & 0.63\sth & 0.60\sth & 0.62\sth \\
    & $\olsBetai{2}$ $(\olsPi{2})$ & -0.21\sth & -0.26\sth & -0.21\sth & -0.18\sth & -0.20\sth \\
    & $\olsBetai{3}$ $(\olsPi{3})$ & -0.16\sth & -0.22\sth & -0.17\sth & -0.14\sth & -0.16\sth \\
    & $\olsBetai{4}$ $(\olsPi{4})$ & -0.08\sth & -0.14\sth & -0.08\stw & -0.05\son & -0.07\stw \\
    & $\olsBetai{5}$ $(\olsPi{5})$ & -0.05\sth & -0.09\sth & -0.03\snn & 0.01\snn & -0.01\snn \\
    & $\olsBetai{6}$ $(\olsPi{6})$ & -0.09\sth & -0.12\sth & -0.07\sth & -0.03\snn & -0.04\son\\
    \bottomrule
  \end{tabular*}
  \caption[Exp. 1: OLS Results. Mixed-Language (Hi), mGTE.]{Estimated OLS coefficients of the positional fairness analysis (similarity between document embedding and standalone segment embeddings) in the mixed-language document setting (\textbf{Hindi}; $\langRemainder=hi$), using mGTE embeddings. $n$ denotes the number of segments per document. OLS regression uses positions as categorical variables; $\olsBetai{0}$ captures the baseline (similarity between document embedding and standalone embedding at position~1), $\olsBetai{p\ge2}$ captures the difference between position $p$'s similarity (similarity between document embedding and standalone embedding at position~$p$) and the baseline similarity. Statistical significance: all values $\mathfrak{p}<0.001$; unless indicated otherwise: \stw $\mathfrak{p}<0.01$, \son $\mathfrak{p}<0.05$, \snn not significant.}
  \label{tab:ols-coef-multi-hi-mgte}
\end{table*}

\begin{table*}[tb]
  \centering
  \small
  \setlength{\tabcolsep}{6pt}
  \begin{tabular*}{\linewidth}{@{\extracolsep{\fill}} ll *{5}{c} @{}}
    \multicolumn{7}{c}{\shortstack{\textbf{OLS Coefficients Positional Fairness}\\ \textbf{(mixed-language Chinese, mGTE)}}} \\
    \addlinespace[2ex]
    \toprule
    \addlinespace[2ex]
    $n$& & \shortstack{[en, zh,\\ \ldots, zh]} & \shortstack{[de, zh,\\ \ldots, zh]} & \shortstack{[it, zh,\\ \ldots, zh]} & \shortstack{[ko, zh,\\ \ldots, zh]} & \shortstack{[hi, zh,\\ \ldots, zh]} \\
    \midrule
    \multirow{3}{*}{3} & $\olsBetai{0}$ $(\olsIntercept)$ & 0.76\sth & 0.81\sth & 0.80\sth & 0.82\sth & 0.83\sth \\
    & $\olsBetai{2}$ $(\olsPi{2})$ & -0.07\sth & -0.19\sth & -0.16\sth & -0.19\sth & -0.20\sth \\
    & $\olsBetai{3}$ $(\olsPi{3})$ & -0.14\sth & -0.22\sth & -0.21\sth & -0.22\sth & -0.23\sth \\
    \addlinespace[2ex]
    \multirow{4}{*}{4} & $\olsBetai{0}$ $(\olsIntercept)$ & 0.72\sth & 0.73\sth & 0.70\sth & 0.75\sth & 0.76\sth \\
    & $\olsBetai{2}$ $(\olsPi{2})$ & -0.05\sth & -0.09\sth & -0.04\sth & -0.12\sth & -0.13\sth \\
    & $\olsBetai{3}$ $(\olsPi{3})$ & -0.10\sth & -0.12\sth & -0.08\sth & -0.13\sth & -0.15\sth \\
    & $\olsBetai{4}$ $(\olsPi{4})$ & -0.11\sth & -0.12\sth & -0.09\sth & -0.14\sth & -0.15\sth \\
    \addlinespace[2ex]
    \multirow{5}{*}{5} & $\olsBetai{0}$ $(\olsIntercept)$ & 0.69\sth & 0.67\sth & 0.63\sth & 0.67\sth & 0.69\sth \\
    & $\olsBetai{2}$ $(\olsPi{2})$ & -0.05\sth & -0.04\sth & 0.01\snn & -0.05\sth & -0.07\sth \\
    & $\olsBetai{3}$ $(\olsPi{3})$ & -0.07\sth & -0.03\stw & 0.01\snn & -0.03\son & -0.05\sth \\
    & $\olsBetai{4}$ $(\olsPi{4})$ & -0.07\sth & -0.03\stw & 0.01\snn & -0.03\son & -0.05\sth \\
    & $\olsBetai{5}$ $(\olsPi{5})$ & -0.09\sth & -0.04\sth & -0.00\snn & -0.04\stw & -0.06\sth \\
    \addlinespace[2ex]
    \multirow{6}{*}{6} & $\olsBetai{0}$ $(\olsIntercept)$ & 0.67\sth & 0.65\sth & 0.60\sth & 0.65\sth & 0.66\sth \\
    & $\olsBetai{2}$ $(\olsPi{2})$ & -0.05\sth & -0.04\sth & 0.02\snn & -0.06\sth & -0.07\sth \\
    & $\olsBetai{3}$ $(\olsPi{3})$ & -0.06\sth & -0.03\sth & 0.02\son & -0.04\sth & -0.04\sth \\
    & $\olsBetai{4}$ $(\olsPi{4})$ & -0.05\sth & -0.01\snn & 0.04\sth & -0.02\son & -0.02\stw \\
    & $\olsBetai{5}$ $(\olsPi{5})$ & -0.06\sth & -0.02\son & 0.03\sth & -0.02\son & -0.03\sth \\
    & $\olsBetai{6}$ $(\olsPi{6})$ & -0.08\sth & -0.04\sth & 0.01\snn & -0.04\sth & -0.05\sth \\
    \bottomrule
  \end{tabular*}
  \caption[Exp. 1: OLS Results. Mixed-Language (Zh), mGTE.]{Estimated OLS coefficients of the positional fairness analysis (similarity between document embedding and standalone segment embeddings) in the mixed-language document setting (\textbf{Chinese}; $\langRemainder=zh$), using mGTE embeddings. $n$ denotes the number of segments per document. OLS regression uses positions as categorical variables; $\olsBetai{0}$ captures the baseline (similarity between document embedding and standalone embedding at position~1), $\olsBetai{p\ge2}$ captures the difference between position $p$'s similarity (similarity between document embedding and standalone embedding at position~$p$) and the baseline similarity. Statistical significance: all values $\mathfrak{p}<0.001$; unless indicated otherwise: \stw $\mathfrak{p}<0.01$, \son $\mathfrak{p}<0.05$, \snn not significant.}
  \label{tab:ols-coef-multi-zh-mgte}
\end{table*}

\begin{table*}[]
  \centering
  \small
  \setlength{\tabcolsep}{6pt}
  \begin{tabular*}{\linewidth}{@{\extracolsep{\fill}} ll *{5}{c} @{}}
    \multicolumn{7}{c}{\shortstack{\textbf{OLS Coefficients Positional Fairness}\\ \textbf{(mixed-language English, jina-v3)}}} \\
    \addlinespace[2ex]
    \toprule
    \addlinespace[2ex]
    $n$& & \shortstack{[zh, en,\\ \ldots, en]} & \shortstack{[de, en,\\ \ldots, en]} & \shortstack{[it, en,\\ \ldots, en]} & \shortstack{[ko, en,\\ \ldots, en]} & \shortstack{[hi, en,\\ \ldots, en]} \\
    \midrule
    \multirow{3}{*}{3} & $\olsBetai{0}$ $(\olsIntercept)$ & 0.74\sth & 0.77\sth & 0.73\sth & 0.68\sth & 0.71\sth \\
    & $\olsBetai{2}$ $(\olsPi{2})$ & -0.18\sth & -0.28\sth & -0.18\sth & -0.09\sth & -0.14\sth \\
    & $\olsBetai{3}$ $(\olsPi{3})$ & -0.26\sth & -0.30\sth & -0.25\sth & -0.20\sth & -0.24\sth \\
    \addlinespace[2ex]
    \multirow{4}{*}{4} & $\olsBetai{0}$ $(\olsIntercept)$ & 0.63\sth & 0.65\sth & 0.60\sth & 0.56\sth & 0.59\sth \\
    & $\olsBetai{2}$ $(\olsPi{2})$ & -0.08\sth & -0.16\sth & -0.06\sth & 0.01\snn & -0.04\sth \\
    & $\olsBetai{3}$ $(\olsPi{3})$ & -0.15\sth & -0.17\sth & -0.12\sth & -0.08\sth & -0.11\sth \\
    & $\olsBetai{4}$ $(\olsPi{4})$ & -0.19\sth & -0.21\sth & -0.16\sth & -0.12\sth & -0.15\sth \\
    \addlinespace[2ex]
    \multirow{5}{*}{5} & $\olsBetai{0}$ $(\olsIntercept)$ & 0.53\sth & 0.53\sth & 0.48\sth & 0.43\sth & 0.48\sth \\
    & $\olsBetai{2}$ $(\olsPi{2})$ & -0.01\snn & -0.05\stw & 0.03\snn & 0.10\sth & 0.04\son\\
    & $\olsBetai{3}$ $(\olsPi{3})$ & -0.04\stw & -0.04\son & 0.02\snn & 0.07\sth & 0.02\snn \\
    & $\olsBetai{4}$ $(\olsPi{4})$ & -0.06\sth & -0.06\sth & -0.01\snn & 0.04\son & -0.01\snn \\
    & $\olsBetai{5}$ $(\olsPi{5})$ & -0.11\sth & -0.12\sth & -0.06\sth & -0.01\snn & -0.06\sth \\
    \addlinespace[2ex]
    \multirow{6}{*}{6} & $\olsBetai{0}$ $(\olsIntercept)$ & 0.50\sth & 0.49\sth & 0.45\sth & 0.40\sth & 0.44\sth \\
    & $\olsBetai{2}$ $(\olsPi{2})$ & -0.01\snn & -0.04\son & 0.02\snn & 0.10\sth & 0.05\sth \\
    & $\olsBetai{3}$ $(\olsPi{3})$ & -0.04\sth & -0.03\stw & 0.01\snn & 0.06\sth & 0.02\son\\
    & $\olsBetai{4}$ $(\olsPi{4})$ & -0.04\sth & -0.04\stw & 0.01\snn & 0.05\sth & 0.02\snn \\
    & $\olsBetai{5}$ $(\olsPi{5})$ & -0.06\sth & -0.07\sth & -0.02\snn & 0.03\son & -0.01\snn \\
    & $\olsBetai{6}$ $(\olsPi{6})$ & -0.11\sth & -0.11\sth & -0.07\sth & -0.02\snn & -0.05\sth \\
    \bottomrule
  \end{tabular*}
  \caption[Exp. 1: OLS Results. Mixed-Language (En), jina-v3.]{Estimated OLS coefficients of the positional fairness analysis (similarity between document embedding and standalone segment embeddings) in the mixed-language document setting (\textbf{English}; $\langRemainder=en$), using jina-v3 embeddings. $n$ denotes the number of segments per document. OLS regression uses positions as categorical variables; $\olsBetai{0}$ captures the baseline (similarity between document embedding and standalone embedding at position~1), $\olsBetai{p\ge2}$ captures the difference between position $p$'s similarity (similarity between document embedding and standalone embedding at position~$p$) and the baseline similarity. Statistical significance: all values $\mathfrak{p}<0.001$; unless indicated otherwise: \stw $\mathfrak{p}<0.01$, \son $\mathfrak{p}<0.05$, \snn not significant.}
  \label{tab:ols-coef-multi-en-jina}
\end{table*}

\begin{table*}[]
  \centering
  \small
  \setlength{\tabcolsep}{6pt}
  \begin{tabular*}{\linewidth}{@{\extracolsep{\fill}} ll *{5}{c} @{}}
    \multicolumn{7}{c}{\shortstack{\textbf{OLS Coefficients Positional Fairness}\\ \textbf{(mixed-language German, jina-v3)}}} \\
    \addlinespace[2ex]
    \toprule
    \addlinespace[2ex]
    $n$& & \shortstack{[en, de,\\ \ldots, de]} & \shortstack{[zh, de,\\ \ldots, de]} & \shortstack{[it, de,\\ \ldots, de]} & \shortstack{[ko, de,\\ \ldots, de]} & \shortstack{[hi, de,\\ \ldots, de]} \\
    \midrule
    \multirow{3}{*}{3} & $\olsBetai{0}$ $(\olsIntercept)$ & 0.87\sth & 0.83\sth & 0.82\sth & 0.79\sth & 0.81\sth \\
    & $\olsBetai{2}$ $(\olsPi{2})$ & -0.44\sth & -0.36\sth & -0.35\sth & -0.28\sth & -0.32\sth \\
    & $\olsBetai{3}$ $(\olsPi{3})$ & -0.53\sth & -0.47\sth & -0.46\sth & -0.42\sth & -0.45\sth \\
    \addlinespace[2ex]
    \multirow{4}{*}{4} & $\olsBetai{0}$ $(\olsIntercept)$ & 0.81\sth & 0.79\sth & 0.76\sth & 0.74\sth & 0.75\sth \\
    & $\olsBetai{2}$ $(\olsPi{2})$ & -0.40\sth & -0.34\sth & -0.31\sth & -0.26\sth & -0.29\sth \\
    & $\olsBetai{3}$ $(\olsPi{3})$ & -0.47\sth & -0.44\sth & -0.41\sth & -0.39\sth & -0.41\sth \\
    & $\olsBetai{4}$ $(\olsPi{4})$ & -0.50\sth & -0.46\sth & -0.43\sth & -0.42\sth & -0.44\sth \\
    \addlinespace[2ex]
    \multirow{5}{*}{5} & $\olsBetai{0}$ $(\olsIntercept)$ & 0.75\sth & 0.74\sth & 0.69\sth & 0.67\sth & 0.69\sth \\
    & $\olsBetai{2}$ $(\olsPi{2})$ & -0.33\sth & -0.29\sth & -0.25\sth & -0.18\sth & -0.22\sth \\
    & $\olsBetai{3}$ $(\olsPi{3})$ & -0.39\sth & -0.38\sth & -0.32\sth & -0.30\sth & -0.33\sth \\
    & $\olsBetai{4}$ $(\olsPi{4})$ & -0.41\sth & -0.40\sth & -0.35\sth & -0.32\sth & -0.36\sth \\
    & $\olsBetai{5}$ $(\olsPi{5})$ & -0.44\sth & -0.42\sth & -0.37\sth & -0.35\sth & -0.38\sth \\
    \addlinespace[2ex]
    \multirow{6}{*}{6} & $\olsBetai{0}$ $(\olsIntercept)$ & 0.66\sth & 0.65\sth & 0.60\sth & 0.58\sth & 0.59\sth \\
    & $\olsBetai{2}$ $(\olsPi{2})$ & -0.26\sth & -0.21\sth & -0.17\sth & -0.12\sth & -0.14\sth \\
    & $\olsBetai{3}$ $(\olsPi{3})$ & -0.28\sth & -0.28\sth & -0.22\sth & -0.20\sth & -0.22\sth \\
    & $\olsBetai{4}$ $(\olsPi{4})$ & -0.30\sth & -0.29\sth & -0.23\sth & -0.21\sth & -0.23\sth \\
    & $\olsBetai{5}$ $(\olsPi{5})$ & -0.31\sth & -0.30\sth & -0.25\sth & -0.23\sth & -0.25\sth \\
    & $\olsBetai{6}$ $(\olsPi{6})$ & -0.34\sth & -0.32\sth & -0.27\sth & -0.25\sth & -0.27\sth \\
    \bottomrule
  \end{tabular*}
  \caption[Exp. 1: OLS Results. Mixed-Language (De), jina-v3.]{Estimated OLS coefficients of the positional fairness analysis (similarity between document embedding and standalone segment embeddings) in the mixed-language document setting (\textbf{German}; $\langRemainder=de$), using jina-v3 embeddings. $n$ denotes the number of segments per document. OLS regression uses positions as categorical variables; $\olsBetai{0}$ captures the baseline (similarity between document embedding and standalone embedding at position~1), $\olsBetai{p\ge2}$ captures the difference between position $p$'s similarity (similarity between document embedding and standalone embedding at position~$p$) and the baseline similarity. Statistical significance: all values $\mathfrak{p}<0.001$.}
  \label{tab:ols-coef-multi-de-jina}
\end{table*}

\begin{table*}[]
  \centering
  \small
  \setlength{\tabcolsep}{6pt}
  \begin{tabular*}{\linewidth}{@{\extracolsep{\fill}} ll *{5}{c} @{}}
    \multicolumn{7}{c}{\shortstack{\textbf{OLS Coefficients Positional Fairness}\\ \textbf{(mixed-language Hindi, jina-v3)}}} \\
    \addlinespace[2ex]
    \toprule
    \addlinespace[2ex]
    $n$& & \shortstack{[en, hi,\\ \ldots, hi]} & \shortstack{[zh, hi,\\ \ldots, hi]} & \shortstack{[de, hi,\\ \ldots, hi]} & \shortstack{[it, hi,\\ \ldots, hi]} & \shortstack{[ko, hi,\\ \ldots, hi]} \\
    \midrule
    \multirow{3}{*}{3} & $\olsBetai{0}$ $(\olsIntercept)$ & 0.88\sth & 0.83\sth & 0.86\sth & 0.83\sth & 0.80\sth \\
    & $\olsBetai{2}$ $(\olsPi{2})$ & -0.49\sth & -0.37\sth & -0.47\sth & -0.39\sth & -0.30\sth \\
    & $\olsBetai{3}$ $(\olsPi{3})$ & -0.53\sth & -0.44\sth & -0.49\sth & -0.44\sth & -0.39\sth \\
    \addlinespace[2ex]
    \multirow{4}{*}{4} & $\olsBetai{0}$ $(\olsIntercept)$ & 0.82\sth & 0.78\sth & 0.80\sth & 0.75\sth & 0.74\sth \\
    & $\olsBetai{2}$ $(\olsPi{2})$ & -0.44\sth & -0.34\sth & -0.43\sth & -0.33\sth & -0.27\sth \\
    & $\olsBetai{3}$ $(\olsPi{3})$ & -0.47\sth & -0.41\sth & -0.45\sth & -0.39\sth & -0.35\sth \\
    & $\olsBetai{4}$ $(\olsPi{4})$ & -0.48\sth & -0.42\sth & -0.45\sth & -0.40\sth & -0.37\sth \\
    \addlinespace[2ex]
    \multirow{5}{*}{5} & $\olsBetai{0}$ $(\olsIntercept)$ & 0.72\sth & 0.69\sth & 0.70\sth & 0.65\sth & 0.62\sth \\
    & $\olsBetai{2}$ $(\olsPi{2})$ & -0.33\sth & -0.25\sth & -0.31\sth & -0.21\sth & -0.14\sth \\
    & $\olsBetai{3}$ $(\olsPi{3})$ & -0.34\sth & -0.30\sth & -0.31\sth & -0.24\sth & -0.21\sth \\
    & $\olsBetai{4}$ $(\olsPi{4})$ & -0.36\sth & -0.31\sth & -0.33\sth & -0.27\sth & -0.23\sth \\
    & $\olsBetai{5}$ $(\olsPi{5})$ & -0.39\sth & -0.34\sth & -0.35\sth & -0.29\sth & -0.26\sth \\
    \addlinespace[2ex]
    \multirow{6}{*}{6} & $\olsBetai{0}$ $(\olsIntercept)$ & 0.69\sth & 0.66\sth & 0.67\sth & 0.62\sth & 0.61\sth \\
    & $\olsBetai{2}$ $(\olsPi{2})$ & -0.31\sth & -0.22\sth & -0.28\sth & -0.20\sth & -0.15\sth \\
    & $\olsBetai{3}$ $(\olsPi{3})$ & -0.32\sth & -0.27\sth & -0.29\sth & -0.23\sth & -0.20\sth \\
    & $\olsBetai{4}$ $(\olsPi{4})$ & -0.33\sth & -0.28\sth & -0.30\sth & -0.25\sth & -0.22\sth \\
    & $\olsBetai{5}$ $(\olsPi{5})$ & -0.35\sth & -0.30\sth & -0.32\sth & -0.27\sth & -0.25\sth \\
    & $\olsBetai{6}$ $(\olsPi{6})$ & -0.37\sth & -0.32\sth & -0.34\sth & -0.29\sth & -0.27\sth \\
    \bottomrule
  \end{tabular*}
  \caption[Exp. 1: OLS Results. Mixed-Language (Hi), jina-v3.]{Estimated OLS coefficients of the positional fairness analysis (similarity between document embedding and standalone segment embeddings) in the mixed-language document setting (\textbf{Hindi}; $\langRemainder=hi$), using jina-v3 embeddings. $n$ denotes the number of segments per document. OLS regression uses positions as categorical variables; $\olsBetai{0}$ captures the baseline (similarity between document embedding and standalone embedding at position~1), $\olsBetai{p\ge2}$ captures the difference between position $p$'s similarity (similarity between document embedding and standalone embedding at position~$p$) and the baseline similarity. Statistical significance: all values $\mathfrak{p}<0.001$.}
  \label{tab:ols-coef-multi-hi-jina}
\end{table*}

\begin{table*}[]
  \centering
  \small
  \setlength{\tabcolsep}{6pt}
  \begin{tabular*}{\linewidth}{@{\extracolsep{\fill}} ll *{5}{c} @{}}
    \multicolumn{7}{c}{\shortstack{\textbf{OLS Coefficients Positional Fairness}\\ \textbf{(mixed-language Chinese, jina-v3)}}} \\
    \addlinespace[2ex]
    \toprule
    \addlinespace[2ex]
    $n$& & \shortstack{[en, zh,\\ \ldots, zh]} & \shortstack{[de, zh,\\ \ldots, zh]} & \shortstack{[it, zh,\\ \ldots, zh]} & \shortstack{[ko, zh,\\ \ldots, zh]} & \shortstack{[hi, zh,\\ \ldots, zh]} \\
    \midrule
    \multirow{3}{*}{3} & $\olsBetai{0}$ $(\olsIntercept)$ & 0.85\sth & 0.83\sth & 0.78\sth & 0.76\sth & 0.77\sth \\
    & $\olsBetai{2}$ $(\olsPi{2})$ & -0.39\sth & -0.38\sth & -0.27\sth & -0.20\sth & -0.23\sth \\
    & $\olsBetai{3}$ $(\olsPi{3})$ & -0.42\sth & -0.38\sth & -0.32\sth & -0.27\sth & -0.30\sth \\
    \addlinespace[2ex]
    \multirow{4}{*}{4} & $\olsBetai{0}$ $(\olsIntercept)$ & 0.79\sth & 0.76\sth & 0.71\sth & 0.70\sth & 0.71\sth \\
    & $\olsBetai{2}$ $(\olsPi{2})$ & -0.32\sth & -0.31\sth & -0.21\sth & -0.15\sth & -0.19\sth \\
    & $\olsBetai{3}$ $(\olsPi{3})$ & -0.36\sth & -0.32\sth & -0.25\sth & -0.22\sth & -0.24\sth \\
    & $\olsBetai{4}$ $(\olsPi{4})$ & -0.36\sth & -0.31\sth & -0.26\sth & -0.23\sth & -0.25\sth \\
    \addlinespace[2ex]
    \multirow{5}{*}{5} & $\olsBetai{0}$ $(\olsIntercept)$ & 0.69\sth & 0.65\sth & 0.59\sth & 0.58\sth & 0.61\sth \\
    & $\olsBetai{2}$ $(\olsPi{2})$ & -0.20\sth & -0.16\sth & -0.07\stw & -0.03\snn & -0.07\stw \\
    & $\olsBetai{3}$ $(\olsPi{3})$ & -0.22\sth & -0.17\sth & -0.11\sth & -0.08\sth & -0.12\sth \\
    & $\olsBetai{4}$ $(\olsPi{4})$ & -0.23\sth & -0.18\sth & -0.12\sth & -0.09\sth & -0.13\sth \\
    & $\olsBetai{5}$ $(\olsPi{5})$ & -0.24\sth & -0.19\sth & -0.13\sth & -0.10\sth & -0.15\sth \\
    \addlinespace[2ex]
    \multirow{6}{*}{6} & $\olsBetai{0}$ $(\olsIntercept)$ & 0.63\sth & 0.58\sth & 0.54\sth & 0.54\sth & 0.55\sth \\
    & $\olsBetai{2}$ $(\olsPi{2})$ & -0.15\sth & -0.10\sth & -0.03\snn & 0.00\snn & -0.03\snn \\
    & $\olsBetai{3}$ $(\olsPi{3})$ & -0.16\sth & -0.11\sth & -0.06\sth & -0.04\stw & -0.06\sth \\
    & $\olsBetai{4}$ $(\olsPi{4})$ & -0.16\sth & -0.12\sth & -0.07\sth & -0.06\sth & -0.07\sth \\
    & $\olsBetai{5}$ $(\olsPi{5})$ & -0.17\sth & -0.12\sth & -0.07\sth & -0.06\sth & -0.08\sth \\
    & $\olsBetai{6}$ $(\olsPi{6})$ & -0.19\sth & -0.13\sth & -0.08\sth & -0.07\sth & -0.08\sth \\
    \bottomrule
  \end{tabular*}
  \caption[Exp. 1: OLS Results. Mixed-Language (Zh), jina-v3.]{Estimated OLS coefficients of the positional fairness analysis (similarity between document embedding and standalone segment embeddings) in the mixed-language document setting (\textbf{Chinese}; $\langRemainder=zh$), using jina-v3 embeddings. $n$ denotes the number of segments per document. OLS regression uses positions as categorical variables; $\olsBetai{0}$ captures the baseline (similarity between document embedding and standalone embedding at position~1), $\olsBetai{p\ge2}$ captures the difference between position $p$'s similarity (similarity between document embedding and standalone embedding at position~$p$) and the baseline similarity. Statistical significance: all values $\mathfrak{p}<0.001$; unless indicated otherwise: \stw $\mathfrak{p}<0.01$, \snn not significant.}
  \label{tab:ols-coef-multi-zh-jina}
\end{table*}

\begin{table*}[tb]
  \centering
  \small
  \setlength{\tabcolsep}{6pt}
  \begin{tabular*}{\linewidth}{@{\extracolsep{\fill}} ll *{6}{c} @{}}
    \multicolumn{8}{c}{\shortstack{\textbf{OLS Coefficients Information Retention}\\\textbf{(monolingual, jina-v3)}}} \\
    \addlinespace[2ex]
    \toprule
    \addlinespace[2ex]
    $n$& & en & zh & de & it & ko & hi \\
    \midrule
    \multirow{3}{*}{3} & $\olsBetai{0}^{(\tau)}$ $(\olsIntercept)$ & 0.92\sth & 0.93\sth & 0.94\sth & 0.93\sth & 0.92\sth & 0.93\sth \\
    & $\olsBetai{2}^{(\tau)}$ $(\olsPi{2})$ & \cEN{-0.25}\sth & -0.24\sth & \cDEbold{-0.41}\sth & -0.34\sth & -0.28\sth & -0.33\sth \\
    & $\olsBetai{3}^{(\tau)}$ $(\olsPi{3})$ & \cEN{-0.20}\sth & -0.22\sth & \cDEbold{-0.38}\sth & -0.30\sth & -0.26\sth & -0.31\sth \\
    \addlinespace[2ex]
    \multirow{4}{*}{4} & $\olsBetai{0}^{(\tau)}$ $(\olsIntercept)$ & 0.88\sth & 0.91\sth & 0.92\sth & 0.91\sth & 0.90\sth & 0.90\sth \\
    & $\olsBetai{2}^{(\tau)}$ $(\olsPi{2})$ & -0.20\sth & -0.25\sth & -0.41\sth & -0.34\sth & -0.30\sth & -0.34\sth \\
    & $\olsBetai{3}^{(\tau)}$ $(\olsPi{3})$ & -0.19\sth & -0.25\sth & -0.39\sth & -0.34\sth & -0.32\sth & -0.34\sth \\
    & $\olsBetai{4}^{(\tau)}$ $(\olsPi{4})$ & -0.17\sth & -0.20\sth & -0.35\sth & -0.28\sth & -0.26\sth & -0.30\sth \\
    \addlinespace[2ex]
    \multirow{5}{*}{5} & $\olsBetai{0}^{(\tau)}$ $(\olsIntercept)$ & 0.82\sth & 0.87\sth & 0.89\sth & 0.88\sth & 0.88\sth & 0.86\sth \\
    & $\olsBetai{2}^{(\tau)}$ $(\olsPi{2})$ & -0.14\sth & -0.21\sth & -0.38\sth & -0.32\sth & -0.31\sth & -0.28\sth \\
    & $\olsBetai{3}^{(\tau)}$ $(\olsPi{3})$ & -0.11\sth & -0.21\sth & -0.36\sth & -0.33\sth & -0.34\sth & -0.28\sth \\
    & $\olsBetai{4}^{(\tau)}$ $(\olsPi{4})$ & -0.12\sth & -0.19\sth & -0.34\sth & -0.31\sth & -0.31\sth & -0.27\sth \\
    & $\olsBetai{5}^{(\tau)}$ $(\olsPi{5})$ & -0.15\sth & -0.17\sth & -0.32\sth & -0.26\sth & -0.25\sth & -0.25\sth \\
    \addlinespace[2ex]
    \multirow{6}{*}{6} & $\olsBetai{0}^{(\tau)}$ $(\olsIntercept)$ & 0.78\sth & 0.84\sth & 0.85\sth & 0.85\sth & 0.85\sth & 0.85\sth \\
    & $\olsBetai{2}^{(\tau)}$ $(\olsPi{2})$ & \cEN{-0.14}\sth & -0.18\sth & \cDEbold{-0.33}\sth & -0.28\sth & -0.27\sth & -0.29\sth \\
    & $\olsBetai{3}^{(\tau)}$ $(\olsPi{3})$ & \cEN{-0.12}\sth & -0.19\sth & \cDEbold{-0.31}\sth & -0.28\sth & -0.28\sth & -0.29\sth \\
    & $\olsBetai{4}^{(\tau)}$ $(\olsPi{4})$ & \cEN{-0.10}\sth & -0.18\sth & \cDEbold{-0.29}\sth & -0.27\sth & -0.27\sth & -0.28\sth \\
    & $\olsBetai{5}^{(\tau)}$ $(\olsPi{5})$ & \cEN{-0.12}\sth & -0.17\sth & \cDEbold{-0.28}\sth & -0.26\sth & -0.26\sth & -0.28\sth \\
    & $\olsBetai{6}^{(\tau)}$ $(\olsPi{6})$ & \cEN{-0.16}\sth & -0.16\sth & \cDEbold{-0.29}\sth & -0.23\sth & -0.23\sth & -0.27\sth \\
    \bottomrule
  \end{tabular*}
  \caption[Exp. 2: OLS Results. Monolingual, jina-v3.]{Estimated OLS coefficients of the information retention analysis (similarity between a standalone segment embedding and its contextualized embedding within a long document) in the monolingual document setting, using jina-v3 embeddings. $n$ denotes the number of segments per document. OLS regression uses positions as categorical variables; $\olsBetai{0}^{(\tau)}$ captures the baseline information retention (similarity between standalone embedding of position~1 and contextualized embedding of position~1 within the document), $\olsBetai{p\ge2}^{(\tau)}$ captures the difference between position $p$'s information retention (similarity between standalone embedding of position~$p$ and contextualized embedding of position~$p$ within the document) and the baseline information retention. Statistical significance: all values $\mathfrak{p}<0.001$.}
  \label{tab:ols-coef-mono-jina-exp2}
\end{table*}

\begin{table*}[tb]
  \centering
  \small
  \setlength{\tabcolsep}{6pt}
  \begin{tabular*}{\linewidth}{@{\extracolsep{\fill}} ll *{5}{c} @{}}
    \multicolumn{7}{c}{\shortstack{\textbf{OLS Coefficients Information Retention}\\\textbf{(mixed-language English, jina-v3)}}} \\
    \addlinespace[2ex]
    \toprule
    \addlinespace[2ex]
    $n$& & \shortstack{[zh, en,\\ \ldots, en]} & \shortstack{[de, en,\\ \ldots, en]} & \shortstack{[it, en,\\ \ldots, en]} & \shortstack{[ko, en,\\ \ldots, en]} & \shortstack{[hi, en,\\ \ldots, en]} \\
    \midrule
    \multirow{3}{*}{3} & $\olsBetai{0}^{(\tau)}$ $(\olsIntercept)$ & 0.89\sth & 0.91\sth & 0.89\sth & 0.87\sth & 0.88\sth \\
    & $\olsBetai{2}^{(\tau)}$ $(\olsPi{2})$ & -0.20\sth & -0.28\sth & -0.21\sth & -0.14\sth & -0.18\sth \\
    & $\olsBetai{3}^{(\tau)}$ $(\olsPi{3})$ & -0.18\sth & -0.20\sth & -0.17\sth & -0.15\sth & -0.17\sth \\
    \addlinespace[2ex]
    \multirow{4}{*}{4} & $\olsBetai{0}^{(\tau)}$ $(\olsIntercept)$ & 0.84\sth & 0.86\sth & 0.84\sth & 0.81\sth & 0.83\sth \\
    & $\olsBetai{2}^{(\tau)}$ $(\olsPi{2})$ & -0.13\sth & -0.22\sth & -0.15\sth & -0.08\sth & -0.12\sth \\
    & $\olsBetai{3}^{(\tau)}$ $(\olsPi{3})$ & -0.16\sth & -0.18\sth & -0.15\sth & -0.12\sth & -0.14\sth \\
    & $\olsBetai{4}^{(\tau)}$ $(\olsPi{4})$ & -0.13\sth & -0.15\sth & -0.12\sth & -0.09\sth & -0.12\sth \\
    \addlinespace[2ex]
    \multirow{5}{*}{5} & $\olsBetai{0}^{(\tau)}$ $(\olsIntercept)$ & 0.78\sth & 0.80\sth & 0.77\sth & 0.72\sth & 0.76\sth \\
    & $\olsBetai{2}^{(\tau)}$ $(\olsPi{2})$ & -0.07\sth & -0.15\sth & -0.08\sth & 0.01\snn & -0.05\stw \\
    & $\olsBetai{3}^{(\tau)}$ $(\olsPi{3})$ & -0.08\sth & -0.10\sth & -0.06\sth & -0.01\snn & -0.05\sth \\
    & $\olsBetai{4}^{(\tau)}$ $(\olsPi{4})$ & -0.07\sth & -0.10\sth & -0.06\sth & -0.02\snn & -0.06\sth \\
    & $\olsBetai{5}^{(\tau)}$ $(\olsPi{5})$ & -0.09\sth & -0.12\sth & -0.08\sth & -0.03\snn & -0.07\sth \\
    \addlinespace[2ex]
    \multirow{6}{*}{6} & $\olsBetai{0}^{(\tau)}$ $(\olsIntercept)$ & 0.73\sth & 0.76\sth & 0.72\sth & 0.67\sth & 0.71\sth \\
    & $\olsBetai{2}^{(\tau)}$ $(\olsPi{2})$ & -0.04\sth & -0.12\sth & -0.06\sth & 0.04\stw & -0.02\snn \\
    & $\olsBetai{3}^{(\tau)}$ $(\olsPi{3})$ & -0.07\sth & -0.10\sth & -0.06\sth & -0.00\snn & -0.04\sth \\
    & $\olsBetai{4}^{(\tau)}$ $(\olsPi{4})$ & -0.05\sth & -0.08\sth & -0.05\sth & 0.01\snn & -0.03\sth \\
    & $\olsBetai{5}^{(\tau)}$ $(\olsPi{5})$ & -0.06\sth & -0.10\sth & -0.06\sth & -0.00\snn & -0.04\sth \\
    & $\olsBetai{6}^{(\tau)}$ $(\olsPi{6})$ & -0.09\sth & -0.13\sth & -0.08\sth & -0.02\son & -0.06\sth \\
    \bottomrule
  \end{tabular*}
  \caption[Exp. 2: OLS Results. Mixed-Language (En), jina-v3.]{Estimated OLS coefficients of the information retention analysis (similarity between a standalone segment embedding and its contextualized embedding within a long document) in the mixed-language document setting (\textbf{English}; $\langRemainder=en$), using jina-v3 embeddings. $n$ denotes the number of segments per document. OLS regression uses positions as categorical variables; $\olsBetai{0}^{(\tau)}$ captures the baseline information retention (similarity between standalone embedding of position~1 and contextualized embedding of position~1 within the document), $\olsBetai{p\ge2}^{(\tau)}$ captures the difference between position $p$'s information retention (similarity between standalone embedding of position~$p$ and contextualized embedding of position~$p$ within the document) and the baseline information retention. Statistical significance: all values $\mathfrak{p}<0.001$; unless indicated otherwise: \stw $\mathfrak{p}<0.01$, \son $\mathfrak{p}<0.05$, \snn not significant.}
  \label{tab:ols-coef-multi-en-jina-exp2}
\end{table*}

\begin{table*}[tb]
  \centering
  \small
  \setlength{\tabcolsep}{6pt}
  \begin{tabular*}{\linewidth}{@{\extracolsep{\fill}} ll *{5}{c} @{}}
    \multicolumn{7}{c}{\shortstack{\textbf{OLS Coefficients Information Retention}\\\textbf{(mixed-language German, jina-v3)}}} \\
    \addlinespace[2ex]
    \toprule
    \addlinespace[2ex]
    $n$& & \shortstack{[en, de,\\ \ldots, de]} & \shortstack{[zh, de,\\ \ldots, de]} & \shortstack{[it, de,\\ \ldots, de]} & \shortstack{[ko, de,\\ \ldots, de]} & \shortstack{[hi, de,\\ \ldots, de]} \\
    \midrule
    \multirow{3}{*}{3} & $\olsBetai{0}^{(\tau)}$ $(\olsIntercept)$ & 0.94\sth & 0.92\sth & 0.92\sth & 0.91\sth & 0.91\sth \\
    & $\olsBetai{2}^{(\tau)}$ $(\olsPi{2})$ & -0.35\sth & -0.31\sth & -0.31\sth & -0.25\sth & -0.28\sth \\
    & $\olsBetai{3}^{(\tau)}$ $(\olsPi{3})$ & -0.38\sth & -0.36\sth & -0.35\sth & -0.34\sth & -0.35\sth \\
    \addlinespace[2ex]
    \multirow{4}{*}{4} & $\olsBetai{0}^{(\tau)}$ $(\olsIntercept)$ & 0.92\sth & 0.90\sth & 0.89\sth & 0.88\sth & 0.88\sth \\
    & $\olsBetai{2}^{(\tau)}$ $(\olsPi{2})$ & -0.35\sth & -0.31\sth & -0.31\sth & -0.26\sth & -0.28\sth \\
    & $\olsBetai{3}^{(\tau)}$ $(\olsPi{3})$ & -0.38\sth & -0.38\sth & -0.35\sth & -0.35\sth & -0.36\sth \\
    & $\olsBetai{4}^{(\tau)}$ $(\olsPi{4})$ & -0.35\sth & -0.33\sth & -0.32\sth & -0.31\sth & -0.32\sth \\
    \addlinespace[2ex]
    \multirow{5}{*}{5} & $\olsBetai{0}^{(\tau)}$ $(\olsIntercept)$ & 0.89\sth & 0.87\sth & 0.86\sth & 0.85\sth & 0.86\sth \\
    & $\olsBetai{2}^{(\tau)}$ $(\olsPi{2})$ & -0.31\sth & -0.29\sth & -0.29\sth & -0.22\sth & -0.25\sth \\
    & $\olsBetai{3}^{(\tau)}$ $(\olsPi{3})$ & -0.35\sth & -0.35\sth & -0.32\sth & -0.31\sth & -0.33\sth \\
    & $\olsBetai{4}^{(\tau)}$ $(\olsPi{4})$ & -0.34\sth & -0.33\sth & -0.31\sth & -0.29\sth & -0.31\sth \\
    & $\olsBetai{5}^{(\tau)}$ $(\olsPi{5})$ & -0.33\sth & -0.31\sth & -0.29\sth & -0.27\sth & -0.29\sth \\
    \addlinespace[2ex]
    \multirow{6}{*}{6} & $\olsBetai{0}^{(\tau)}$ $(\olsIntercept)$ & 0.85\sth & 0.82\sth & 0.81\sth & 0.79\sth & 0.79\sth \\
    & $\olsBetai{2}^{(\tau)}$ $(\olsPi{2})$ & -0.29\sth & -0.24\sth & -0.23\sth & -0.18\sth & -0.19\sth \\
    & $\olsBetai{3}^{(\tau)}$ $(\olsPi{3})$ & -0.30\sth & -0.29\sth & -0.26\sth & -0.25\sth & -0.26\sth \\
    & $\olsBetai{4}^{(\tau)}$ $(\olsPi{4})$ & -0.28\sth & -0.27\sth & -0.24\sth & -0.23\sth & -0.24\sth \\
    & $\olsBetai{5}^{(\tau)}$ $(\olsPi{5})$ & -0.28\sth & -0.26\sth & -0.24\sth & -0.23\sth & -0.23\sth \\
    & $\olsBetai{6}^{(\tau)}$ $(\olsPi{6})$ & -0.29\sth & -0.26\sth & -0.24\sth & -0.23\sth & -0.23\sth \\
    \bottomrule
  \end{tabular*}
  \caption[Exp. 2: OLS Results. Mixed-Language (De), jina-v3.]{Estimated OLS coefficients of the information retention analysis (similarity between a standalone segment embedding and its contextualized embedding within a long document) in the mixed-language document setting (\textbf{German}; $\langRemainder=de$), using jina-v3 embeddings. $n$ denotes the number of segments per document. OLS regression uses positions as categorical variables; $\olsBetai{0}^{(\tau)}$ captures the baseline information retention (similarity between standalone embedding of position~1 and contextualized embedding of position~1 within the document), $\olsBetai{p\ge2}^{(\tau)}$ captures the difference between position $p$'s information retention (similarity between standalone embedding of position~$p$ and contextualized embedding of position~$p$ within the document) and the baseline information retention. Statistical significance: all values $\mathfrak{p}<0.001$.}
  \label{tab:ols-coef-multi-de-jina-exp2}
\end{table*}

\begin{table*}[tb]
  \centering
  \small
  \setlength{\tabcolsep}{6pt}
  \begin{tabular*}{\linewidth}{@{\extracolsep{\fill}} ll *{5}{c} @{}}
    \multicolumn{7}{c}{\shortstack{\textbf{OLS Coefficients Information Retention}\\\textbf{(mixed-language Hindi, jina-v3)}}} \\
    \addlinespace[2ex]
    \toprule
    \addlinespace[2ex]
    $n$& & \shortstack{[en, hi,\\ \ldots, hi]} & \shortstack{[zh, hi,\\ \ldots, hi]} & \shortstack{[de, hi,\\ \ldots, hi]} & \shortstack{[it, hi,\\ \ldots, hi]} & \shortstack{[ko, hi,\\ \ldots, hi]} \\
    \midrule
    \multirow{3}{*}{3} & $\olsBetai{0}^{(\tau)}$ $(\olsIntercept)$ & 0.95\sth & 0.93\sth & 0.94\sth & 0.93\sth & 0.92\sth \\
    & $\olsBetai{2}^{(\tau)}$ $(\olsPi{2})$ & -0.39\sth & -0.33\sth & -0.42\sth & -0.34\sth & -0.28\sth \\
    & $\olsBetai{3}^{(\tau)}$ $(\olsPi{3})$ & -0.35\sth & -0.32\sth & -0.35\sth & -0.31\sth & -0.30\sth \\
    \addlinespace[2ex]
    \multirow{4}{*}{4} & $\olsBetai{0}^{(\tau)}$ $(\olsIntercept)$ & 0.93\sth & 0.90\sth & 0.92\sth & 0.89\sth & 0.89\sth \\
    & $\olsBetai{2}^{(\tau)}$ $(\olsPi{2})$ & -0.39\sth & -0.33\sth & -0.43\sth & -0.34\sth & -0.28\sth \\
    & $\olsBetai{3}^{(\tau)}$ $(\olsPi{3})$ & -0.38\sth & -0.35\sth & -0.38\sth & -0.34\sth & -0.32\sth \\
    & $\olsBetai{4}^{(\tau)}$ $(\olsPi{4})$ & -0.33\sth & -0.29\sth & -0.32\sth & -0.28\sth & -0.28\sth \\
    \addlinespace[2ex]
    \multirow{5}{*}{5} & $\olsBetai{0}^{(\tau)}$ $(\olsIntercept)$ & 0.89\sth & 0.86\sth & 0.88\sth & 0.85\sth & 0.84\sth \\
    & $\olsBetai{2}^{(\tau)}$ $(\olsPi{2})$ & -0.34\sth & -0.27\sth & -0.37\sth & -0.27\sth & -0.21\sth \\
    & $\olsBetai{3}^{(\tau)}$ $(\olsPi{3})$ & -0.32\sth & -0.29\sth & -0.32\sth & -0.27\sth & -0.25\sth \\
    & $\olsBetai{4}^{(\tau)}$ $(\olsPi{4})$ & -0.30\sth & -0.27\sth & -0.29\sth & -0.25\sth & -0.23\sth \\
    & $\olsBetai{5}^{(\tau)}$ $(\olsPi{5})$ & -0.29\sth & -0.25\sth & -0.27\sth & -0.22\sth & -0.22\sth \\
    \addlinespace[2ex]
    \multirow{6}{*}{6} & $\olsBetai{0}^{(\tau)}$ $(\olsIntercept)$ & 0.88\sth & 0.83\sth & 0.86\sth & 0.82\sth & 0.82\sth \\
    & $\olsBetai{2}^{(\tau)}$ $(\olsPi{2})$ & -0.35\sth & -0.26\sth & -0.35\sth & -0.26\sth & -0.22\sth \\
    & $\olsBetai{3}^{(\tau)}$ $(\olsPi{3})$ & -0.33\sth & -0.28\sth & -0.32\sth & -0.27\sth & -0.26\sth \\
    & $\olsBetai{4}^{(\tau)}$ $(\olsPi{4})$ & -0.31\sth & -0.27\sth & -0.30\sth & -0.26\sth & -0.25\sth \\
    & $\olsBetai{5}^{(\tau)}$ $(\olsPi{5})$ & -0.31\sth & -0.27\sth & -0.29\sth & -0.25\sth & -0.25\sth \\
    & $\olsBetai{6}^{(\tau)}$ $(\olsPi{6})$ & -0.31\sth & -0.25\sth & -0.29\sth & -0.24\sth & -0.24\sth \\
    \bottomrule
  \end{tabular*}
  \caption[Exp. 2: OLS Results. Mixed-Language (Hi), jina-v3.]{Estimated OLS coefficients of the information retention analysis (similarity between a standalone segment embedding and its contextualized embedding within a long document) in the mixed-language document setting (\textbf{Hindi}; $\langRemainder=hi$), using jina-v3 embeddings. $n$ denotes the number of segments per document. OLS regression uses positions as categorical variables; $\olsBetai{0}^{(\tau)}$ captures the baseline information retention (similarity between standalone embedding of position~1 and contextualized embedding of position~1 within the document), $\olsBetai{p\ge2}^{(\tau)}$ captures the difference between position $p$'s information retention (similarity between standalone embedding of position~$p$ and contextualized embedding of position~$p$ within the document) and the baseline information retention. Statistical significance: all values $\mathfrak{p}<0.001$.}
  \label{tab:ols-coef-multi-hi-jina-exp2}
\end{table*}

\begin{table*}[tb]
  \centering
  \small
  \setlength{\tabcolsep}{6pt}
  \begin{tabular*}{\linewidth}{@{\extracolsep{\fill}} ll *{5}{c} @{}}
    \multicolumn{7}{c}{\shortstack{\textbf{OLS Coefficients Information Retention}\\\textbf{(mixed-language Chinese, jina-v3)}}} \\
    \addlinespace[2ex]
    \toprule
    \addlinespace[2ex]
    $n$& & \shortstack{[en, zh,\\ \ldots, zh]} & \shortstack{[de, zh,\\ \ldots, zh]} & \shortstack{[it, zh,\\ \ldots, zh]} & \shortstack{[ko, zh,\\ \ldots, zh]} & \shortstack{[hi, zh,\\ \ldots, zh]} \\
    \midrule
    \multirow{3}{*}{3} & $\olsBetai{0}^{(\tau)}$ $(\olsIntercept)$ & 0.94\sth & 0.92\sth & 0.90\sth & 0.89\sth & 0.90\sth \\
    & $\olsBetai{2}^{(\tau)}$ $(\olsPi{2})$ & -0.30\sth & -0.34\sth & -0.25\sth & -0.20\sth & -0.22\sth \\
    & $\olsBetai{3}^{(\tau)}$ $(\olsPi{3})$ & -0.27\sth & -0.26\sth & -0.22\sth & -0.20\sth & -0.21\sth \\
    \addlinespace[2ex]
    \multirow{4}{*}{4} & $\olsBetai{0}^{(\tau)}$ $(\olsIntercept)$ & 0.91\sth & 0.89\sth & 0.87\sth & 0.86\sth & 0.87\sth \\
    & $\olsBetai{2}^{(\tau)}$ $(\olsPi{2})$ & -0.29\sth & -0.31\sth & -0.23\sth & -0.19\sth & -0.21\sth \\
    & $\olsBetai{3}^{(\tau)}$ $(\olsPi{3})$ & -0.27\sth & -0.26\sth & -0.22\sth & -0.21\sth & -0.22\sth \\
    & $\olsBetai{4}^{(\tau)}$ $(\olsPi{4})$ & -0.22\sth & -0.20\sth & -0.17\sth & -0.16\sth & -0.17\sth \\
    \addlinespace[2ex]
    \multirow{5}{*}{5} & $\olsBetai{0}^{(\tau)}$ $(\olsIntercept)$ & 0.87\sth & 0.84\sth & 0.81\sth & 0.81\sth & 0.82\sth \\
    & $\olsBetai{2}^{(\tau)}$ $(\olsPi{2})$ & -0.22\sth & -0.23\sth & -0.15\sth & -0.11\sth & -0.14\sth \\
    & $\olsBetai{3}^{(\tau)}$ $(\olsPi{3})$ & -0.22\sth & -0.20\sth & -0.16\sth & -0.14\sth & -0.16\sth \\
    & $\olsBetai{4}^{(\tau)}$ $(\olsPi{4})$ & -0.20\sth & -0.17\sth & -0.13\sth & -0.12\sth & -0.14\sth \\
    & $\olsBetai{5}^{(\tau)}$ $(\olsPi{5})$ & -0.18\sth & -0.15\sth & -0.10\sth & -0.09\sth & -0.12\sth \\
    \addlinespace[2ex]
    \multirow{6}{*}{6} & $\olsBetai{0}^{(\tau)}$ $(\olsIntercept)$ & 0.84\sth & 0.80\sth & 0.77\sth & 0.78\sth & 0.78\sth \\
    & $\olsBetai{2}^{(\tau)}$ $(\olsPi{2})$ & -0.21\sth & -0.19\sth & -0.12\sth & -0.10\sth & -0.12\sth \\
    & $\olsBetai{3}^{(\tau)}$ $(\olsPi{3})$ & -0.19\sth & -0.17\sth & -0.12\sth & -0.12\sth & -0.13\sth \\
    & $\olsBetai{4}^{(\tau)}$ $(\olsPi{4})$ & -0.18\sth & -0.15\sth & -0.11\sth & -0.11\sth & -0.12\sth \\
    & $\olsBetai{5}^{(\tau)}$ $(\olsPi{5})$ & -0.17\sth & -0.14\sth & -0.10\sth & -0.10\sth & -0.11\sth \\
    & $\olsBetai{6}^{(\tau)}$ $(\olsPi{6})$ & -0.17\sth & -0.13\sth & -0.09\sth & -0.09\sth & -0.10\sth \\
    \bottomrule
  \end{tabular*}
  \caption[Exp. 2: OLS Results. Mixed-Language (Zh), jina-v3.]{Estimated OLS coefficients of the information retention analysis (similarity between a standalone segment embedding and its contextualized embedding within a long document) in the mixed-language document setting (\textbf{Chinese}; $\langRemainder=zh$), using jina-v3 embeddings. $n$ denotes the number of segments per document. OLS regression uses positions as categorical variables; $\olsBetai{0}^{(\tau)}$ captures the baseline information retention (similarity between standalone embedding of position~1 and contextualized embedding of position~1 within the document), $\olsBetai{p\ge2}^{(\tau)}$ captures the difference between position $p$'s information retention (similarity between standalone embedding of position~$p$ and contextualized embedding of position~$p$ within the document) and the baseline information retention. Statistical significance: all values $\mathfrak{p}<0.001$.}
  \label{tab:ols-coef-multi-zh-jina-exp2}
\end{table*}

\begin{figure*}[tb]
  \centering
  \begin{subfigure}{\textwidth}
    \centering
    \includegraphics[width=\textwidth]{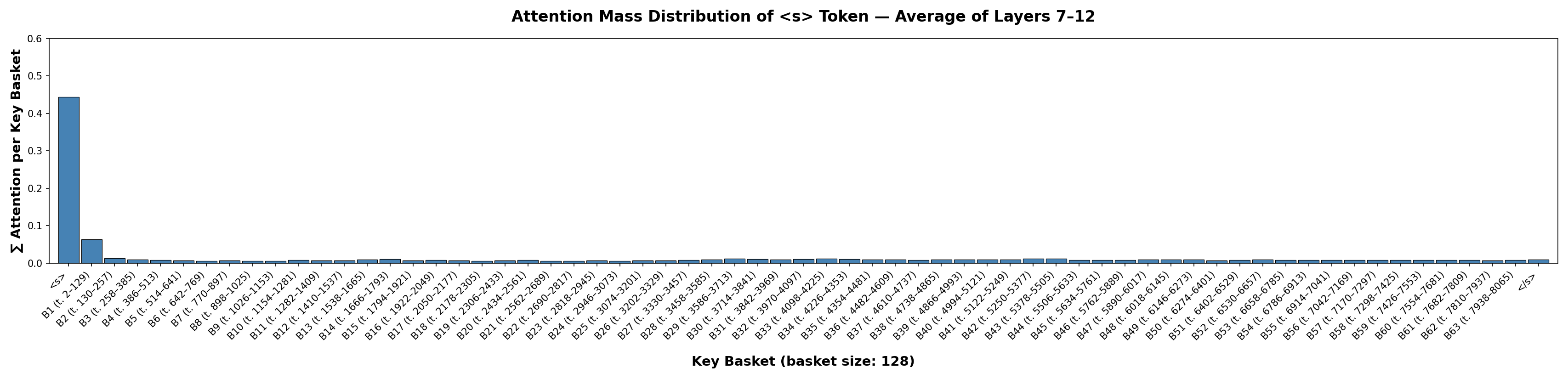}
    \caption{English}
    \label{fig:exp3_self_attn_mono_en_5seg_layerAVG}
  \end{subfigure}
  \par\vspace{2em}
  \begin{subfigure}{\textwidth}
    \centering
    \includegraphics[width=\textwidth]{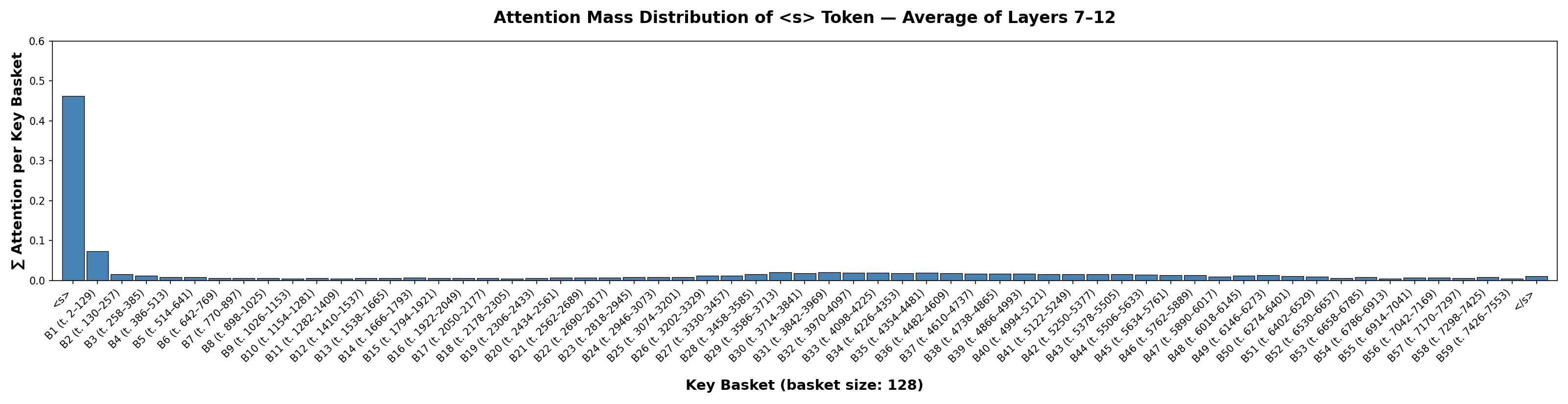}
    \caption{Hindi}
    \label{fig:exp3_self_attn_mono_hi_5seg_layerAVG}
  \end{subfigure}
  \caption{Front-loaded self-attention distribution of the $\poolingToken$-query token over key baskets (basket size $\basketSize{=}128$) in English (top) and Hindi (bottom) documents ($n{=}5$). For Hindi, we additionally observe slight mid/late-sequence increases in the attention distribution, leading to U-shaped attention profiles. Average of the last six transformer layers.}
  \label{fig:exp3_self_attn_mono_enhi_5seg_layerAVG}
\end{figure*}

\begin{figure*}
  \centering
  \includegraphics[width=\textwidth]{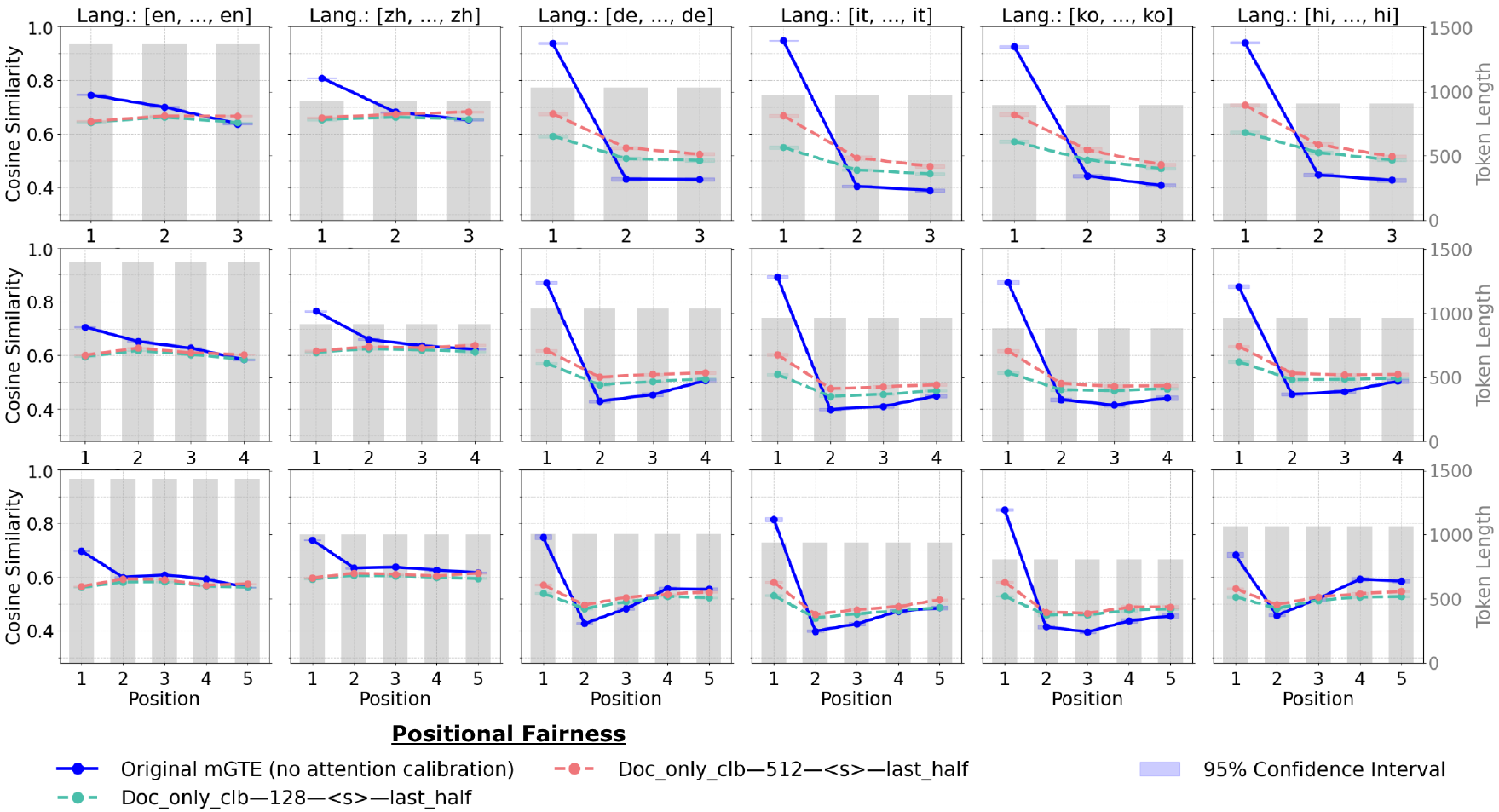}
  \caption{Control experiment to test for semantic fidelity of the attention calibration approach. Both dashed and solid lines use uncalibrated segment embeddings. Solid line uses uncalibrated document embeddings; dashed lines use calibrated document embeddings.}
  \label{fig:EF_mono_calibrated_DocOnly}
\end{figure*}

\begin{table*}[tb]
  \centering
  \small
  \setlength{\tabcolsep}{6pt}
  \begin{tabular*}{\linewidth}{@{\extracolsep{\fill}} ll *{6}{c@{\hspace{2pt}}c} @{}}
    \multicolumn{14}{c}{\shortstack{\textbf{OLS Coefficients Positional Fairness}\ \textbf{(monolingual, mGTE, calibration effect)}}} \\
    \addlinespace[2ex]
    \toprule
    \addlinespace[2ex]
    $n$ & & \multicolumn{2}{c}{en} & \multicolumn{2}{c}{zh} & \multicolumn{2}{c}{de} & \multicolumn{2}{c}{it} & \multicolumn{2}{c}{ko} & \multicolumn{2}{c}{hi} \\
    \cmidrule(lr){3-4} \cmidrule(lr){5-6} \cmidrule(lr){7-8} \cmidrule(lr){9-10} \cmidrule(lr){11-12} \cmidrule(lr){13-14}
    & & c. & \textit{uc.} & c. & \textit{uc.} & c. & \textit{uc.} & c. & \textit{uc.} & c. & \textit{uc.} & c. & \textit{uc.} \\
    \midrule
    \multirow{3}{*}{3} & $\olsBetai{0}$ & 0.70 & 0.75 & 0.71 & 0.81 & 0.68 & 0.94 & 0.64 & 0.95 & 0.69 & 0.93 & 0.69 & 0.94 \\
    & $\olsBetai{2}$ & \cEN{\underline{0.02}} & -0.05 & 0.01 & -0.13 & \cDE{\underline{-0.08}} & -0.51 & -0.09 & -0.54 & -0.07 & -0.48 & -0.08 & -0.49 \\
    & $\olsBetai{3}$ & \cEN{\underline{-0.00}} & -0.11 & -0.00 & -0.16 & \cDE{\underline{-0.09}} & -0.51 & -0.10 & -0.56 & -0.11 & -0.52 & -0.10 & -0.51 \\
    \addlinespace[2ex]
    \multirow{4}{*}{4} & $\olsBetai{0}$ & 0.65 & 0.71 & 0.67 & 0.77 & 0.65 & 0.87 & 0.61 & 0.89 & 0.64 & 0.87 & 0.66 & 0.86 \\
    & $\olsBetai{2}$ & 0.02 & -0.05 & 0.01 & -0.11 & -0.07 & -0.44 & -0.08 & -0.49 & -0.07 & -0.44 & -0.07 & -0.40 \\
    & $\olsBetai{3}$ & 0.01 & -0.08 & 0.01 & -0.13 & -0.07 & -0.42 & -0.07 & -0.48 & -0.07 & -0.46 & -0.07 & -0.39 \\
    & $\olsBetai{4}$ & -0.01 & -0.12 & -0.00 & -0.14 & -0.06 & -0.36 & -0.06 & -0.44 & -0.06 & -0.43 & -0.06 & -0.35 \\
    \addlinespace[2ex]
    \multirow{5}{*}{5} & $\olsBetai{0}$ & 0.61 & 0.70 & 0.65 & 0.74 & 0.61 & 0.75 & 0.61 & 0.82 & 0.61 & 0.85 & 0.61 & 0.68 \\
    & $\olsBetai{2}$ & 0.02 & -0.10 & 0.01 & -0.10 & -0.05 & -0.32 & -0.08 & -0.42 & -0.07 & -0.44 & -0.04 & -0.22 \\
    & $\olsBetai{3}$ & 0.02 & -0.09 & 0.01 & -0.10 & -0.03 & -0.27 & -0.07 & -0.39 & -0.07 & -0.46 & -0.01 & -0.16 \\
    & $\olsBetai{4}$ & 0.01 & -0.10 & 0.01 & -0.11 & -0.01 & -0.19 & -0.06 & -0.34 & -0.05 & -0.41 & -0.00 & -0.09 \\
    & $\olsBetai{5}$ & 0.00 & -0.14 & 0.00 & -0.12 & -0.02 & -0.19 & -0.04 & -0.33 & -0.05 & -0.40 & 0.00 & -0.10 \\
    \bottomrule
  \end{tabular*}
  \caption[Exp. 1: OLS Results. Monolingual, Calibrated.]{Comparison between attention-calibrated (c.) and uncalibrated (\textit{uc.}) mGTE embeddings. Estimated OLS coefficients of the positional fairness analysis (similarity between document embedding and standalone segment embeddings) in the monolingual document setting. $n$ denotes the number of segments per document. OLS regression uses positions as categorical variables; $\olsBetai{0}$ captures the baseline (similarity between document embedding and standalone embedding at position~1), $\olsBetai{p\ge2}$ captures the difference between position $p$'s similarity (similarity between document embedding and standalone embedding at position~$p$) and the baseline similarity. Calibrated embeddings use the following hyperparameters: $\basketSize=128$, $\layersCal=\{7,\ldots,12\}$.}
  \label{tab:ols-coef-mono-alibaba-nlp-gte-multilingual-base-calibrated-comparison}
\end{table*}

\begin{table*}[tb]
  \centering
  \small
  \setlength{\tabcolsep}{6pt}
  \begin{tabular*}{\linewidth}{@{\extracolsep{\fill}} ll *{5}{c@{\hspace{2pt}}c} @{}}
    \multicolumn{12}{c}{\shortstack{\textbf{OLS Coefficients Positional Fairness}\\ \textbf{(mixed-language English, mGTE, calibrated)}}} \\
    \addlinespace[2ex]
    \toprule
    \addlinespace[2ex]
    $n$& & \multicolumn{2}{c}{\shortstack{[zh, en,\\ \ldots, en]}} & \multicolumn{2}{c}{\shortstack{[de, en,\\ \ldots, en]}} & \multicolumn{2}{c}{\shortstack{[it, en,\\ \ldots, en]}} & \multicolumn{2}{c}{\shortstack{[ko, en,\\ \ldots, en]}} & \multicolumn{2}{c}{\shortstack{[hi, en,\\ \ldots, en]}} \\
    \cmidrule(lr){3-4} \cmidrule(lr){5-6} \cmidrule(lr){7-8} \cmidrule(lr){9-10} \cmidrule(lr){11-12}
    & & c. & \textit{uc.} & c. & \textit{uc.} & c. & \textit{uc.} & c. & \textit{uc.} & c. & \textit{uc.} \\
    \midrule
    \multirow{3}{*}{3} & $\olsBetai{0}$ & 0.65 & 0.77 & 0.62 & 0.72 & 0.60 & 0.71 & 0.60 & 0.72 & 0.63 & 0.72 \\
    & $\olsBetai{2}$ & 0.08 & -0.09 & 0.11 & -0.03 & 0.13 & -0.02 & 0.14 & -0.03 & 0.11 & -0.03 \\
    & $\olsBetai{3}$ & 0.07 & -0.13 & 0.09 & -0.06 & 0.11 & -0.05 & 0.12 & -0.05 & 0.09 & -0.05 \\
    \addlinespace[2ex]
    \multirow{4}{*}{4} & $\olsBetai{0}$ & 0.61 & 0.73 & 0.58 & 0.67 & 0.56 & 0.64 & 0.57 & 0.65 & 0.59 & 0.67 \\
    & $\olsBetai{2}$ & 0.07 & -0.10 & 0.10 & -0.03 & 0.13 & -0.01 & 0.12 & -0.01 & 0.09 & -0.02 \\
    & $\olsBetai{3}$ & 0.06 & -0.10 & 0.10 & -0.01 & 0.12 & 0.02 & 0.11 & 0.01 & 0.08 & -0.01 \\
    & $\olsBetai{4}$ & 0.05 & -0.14 & 0.08 & -0.05 & 0.10 & -0.02 & 0.10 & -0.03 & 0.07 & -0.05 \\
    \addlinespace[2ex]
    \multirow{5}{*}{5} & $\olsBetai{0}$ & 0.59 & 0.71 & 0.54 & 0.65 & 0.53 & 0.62 & 0.53 & 0.62 & 0.57 & 0.65 \\
    & $\olsBetai{2}$ & 0.05 & -0.12 & 0.10 & -0.07 & 0.12 & -0.02 & 0.12 & -0.03 & 0.08 & -0.06 \\
    & $\olsBetai{3}$ & 0.06 & -0.10 & 0.11 & -0.03 & 0.12 & 0.02 & 0.12 & 0.01 & 0.08 & -0.02 \\
    & $\olsBetai{4}$ & 0.04 & -0.11 & 0.10 & -0.03 & 0.11 & 0.01 & 0.11 & 0.01 & 0.07 & -0.02 \\
    & $\olsBetai{5}$ & 0.04 & -0.14 & 0.09 & -0.07 & 0.10 & -0.03 & 0.10 & -0.03 & 0.06 & -0.07 \\
    \bottomrule
  \end{tabular*}
  \caption[Exp. 1: OLS Results. Mixed-Language (En), Calibrated.]{Comparison between attention-calibrated (c.) and uncalibrated (\textit{uc.}) mGTE embeddings. Estimated OLS coefficients of the positional fairness analysis (similarity between document embedding and standalone segment embeddings) in the mixed-language document setting (\textbf{English}; $\langRemainder=en$). $n$ denotes the number of segments per document. OLS regression uses positions as categorical variables; $\olsBetai{0}$ captures the baseline (similarity between document embedding and standalone embedding at position~1), $\olsBetai{p\ge2}$ captures the difference between position $p$'s similarity (similarity between document embedding and standalone embedding at position~$p$) and the baseline similarity. Calibrated embeddings use the following hyperparameters: $\basketSize=128$, $\layersCal=\{7,\ldots,12\}$.}
  \label{tab:ols-coef-multi-en-mgte-calibrated-comparison}
\end{table*}

\begin{table*}[tb]
  \centering
  \small
  \setlength{\tabcolsep}{6pt}
  \begin{tabular*}{\linewidth}{@{\extracolsep{\fill}} ll *{5}{c@{\hspace{2pt}}c} @{}}
    \multicolumn{12}{c}{\shortstack{\textbf{OLS Coefficients Positional Fairness}\\ \textbf{(mixed-language German, mGTE, calibrated)}}} \\
    \addlinespace[2ex]
    \toprule
    \addlinespace[2ex]
    $n$& & \multicolumn{2}{c}{\shortstack{[en, de,\\ \ldots, de]}} & \multicolumn{2}{c}{\shortstack{[zh, de,\\ \ldots, de]}} & \multicolumn{2}{c}{\shortstack{[it, de,\\ \ldots, de]}} & \multicolumn{2}{c}{\shortstack{[ko, de,\\ \ldots, de]}} & \multicolumn{2}{c}{\shortstack{[hi, de,\\ \ldots, de]}} \\
    \cmidrule(lr){3-4} \cmidrule(lr){5-6} \cmidrule(lr){7-8} \cmidrule(lr){9-10} \cmidrule(lr){11-12}
    & & c. & \textit{uc.} & c. & \textit{uc.} & c. & \textit{uc.} & c. & \textit{uc.} & c. & \textit{uc.} \\
    \midrule
    \multirow{3}{*}{3} & $\olsBetai{0}$ & 0.74 & 0.83 & 0.69 & 0.89 & 0.67 & 0.92 & 0.67 & 0.91 & 0.70 & 0.92 \\
    & $\olsBetai{2}$ & -0.09 & -0.26 & -0.03 & -0.36 & -0.06 & -0.48 & -0.04 & -0.45 & -0.07 & -0.45 \\
    & $\olsBetai{3}$ & -0.11 & -0.29 & -0.05 & -0.38 & -0.08 & -0.49 & -0.08 & -0.48 & -0.10 & -0.47 \\
    \addlinespace[2ex]
    \multirow{4}{*}{4} & $\olsBetai{0}$ & 0.68 & 0.77 & 0.65 & 0.84 & 0.63 & 0.84 & 0.64 & 0.85 & 0.66 & 0.83 \\
    & $\olsBetai{2}$ & -0.07 & -0.24 & -0.04 & -0.36 & -0.05 & -0.40 & -0.04 & -0.38 & -0.06 & -0.36 \\
    & $\olsBetai{3}$ & -0.07 & -0.23 & -0.04 & -0.34 & -0.05 & -0.38 & -0.05 & -0.39 & -0.07 & -0.35 \\
    & $\olsBetai{4}$ & -0.08 & -0.22 & -0.05 & -0.33 & -0.04 & -0.34 & -0.04 & -0.35 & -0.06 & -0.31 \\
    \addlinespace[2ex]
    \multirow{5}{*}{5} & $\olsBetai{0}$ & 0.64 & 0.69 & 0.63 & 0.78 & 0.59 & 0.73 & 0.60 & 0.74 & 0.62 & 0.69 \\
    & $\olsBetai{2}$ & -0.07 & -0.21 & -0.05 & -0.32 & -0.04 & -0.30 & -0.03 & -0.30 & -0.06 & -0.24 \\
    & $\olsBetai{3}$ & -0.05 & -0.16 & -0.04 & -0.28 & -0.02 & -0.25 & -0.02 & -0.28 & -0.04 & -0.19 \\
    & $\olsBetai{4}$ & -0.05 & -0.11 & -0.03 & -0.23 & 0.00 & -0.18 & 0.00 & -0.20 & -0.02 & -0.11 \\
    & $\olsBetai{5}$ & -0.06 & -0.14 & -0.04 & -0.23 & -0.00 & -0.18 & 0.00 & -0.19 & -0.03 & -0.12 \\
    \bottomrule
  \end{tabular*}
  \caption[Exp. 1: OLS Results. Mixed-Language (De), Calibrated.]{Comparison between attention-calibrated (c.) and uncalibrated (\textit{uc.}) mGTE embeddings. Estimated OLS coefficients of the positional fairness analysis (similarity between document embedding and standalone segment embeddings) in the mixed-language document setting (\textbf{German}; $\langRemainder=de$). $n$ denotes the number of segments per document. OLS regression uses positions as categorical variables; $\olsBetai{0}$ captures the baseline (similarity between document embedding and standalone embedding at position~1), $\olsBetai{p\ge2}$ captures the difference between position $p$'s similarity (similarity between document embedding and standalone embedding at position~$p$) and the baseline similarity. Calibrated embeddings use the following hyperparameters: $\basketSize=128$, $\layersCal=\{7,\ldots,12\}$.}
  \label{tab:ols-coef-multi-de-mgte-calibrated-comparison}
\end{table*}

\begin{table*}[tb]
  \centering
  \small
  \setlength{\tabcolsep}{6pt}
  \begin{tabular*}{\linewidth}{@{\extracolsep{\fill}} ll *{5}{c@{\hspace{2pt}}c} @{}}
    \multicolumn{12}{c}{\shortstack{\textbf{OLS Coefficients Positional Fairness}\\ \textbf{(mixed-language Chinese, mGTE, calibrated)}}} \\
    \addlinespace[2ex]
    \toprule
    \addlinespace[2ex]
    $n$& & \multicolumn{2}{c}{\shortstack{[en, zh,\\ \ldots, zh]}} & \multicolumn{2}{c}{\shortstack{[de, zh,\\ \ldots, zh]}} & \multicolumn{2}{c}{\shortstack{[it, zh,\\ \ldots, zh]}} & \multicolumn{2}{c}{\shortstack{[ko, zh,\\ \ldots, zh]}} & \multicolumn{2}{c}{\shortstack{[hi, zh,\\ \ldots, zh]}} \\
    \cmidrule(lr){3-4} \cmidrule(lr){5-6} \cmidrule(lr){7-8} \cmidrule(lr){9-10} \cmidrule(lr){11-12}
    & & c. & \textit{uc.} & c. & \textit{uc.} & c. & \textit{uc.} & c. & \textit{uc.} & c. & \textit{uc.} \\
    \midrule
    \multirow{3}{*}{3} & $\olsBetai{0}$ & 0.74 & 0.76 & 0.67 & 0.81 & 0.66 & 0.80 & 0.66 & 0.82 & 0.68 & 0.83 \\
    & $\olsBetai{2}$ & -0.05 & -0.07 & 0.01 & -0.19 & 0.03 & -0.16 & 0.04 & -0.19 & 0.01 & -0.20 \\
    & $\olsBetai{3}$ & -0.06 & -0.14 & -0.00 & -0.22 & 0.01 & -0.21 & 0.02 & -0.22 & -0.00 & -0.23 \\
    \addlinespace[2ex]
    \multirow{4}{*}{4} & $\olsBetai{0}$ & 0.69 & 0.72 & 0.62 & 0.73 & 0.60 & 0.70 & 0.62 & 0.75 & 0.65 & 0.76 \\
    & $\olsBetai{2}$ & -0.03 & -0.05 & 0.05 & -0.09 & 0.07 & -0.04 & 0.05 & -0.12 & 0.02 & -0.13 \\
    & $\olsBetai{3}$ & -0.04 & -0.10 & 0.04 & -0.12 & 0.07 & -0.08 & 0.04 & -0.13 & 0.01 & -0.15 \\
    & $\olsBetai{4}$ & -0.05 & -0.11 & 0.03 & -0.12 & 0.06 & -0.09 & 0.04 & -0.14 & 0.01 & -0.15 \\
    \addlinespace[2ex]
    \multirow{5}{*}{5} & $\olsBetai{0}$ & 0.66 & 0.69 & 0.58 & 0.67 & 0.56 & 0.63 & 0.58 & 0.67 & 0.61 & 0.69 \\
    & $\olsBetai{2}$ & -0.02 & -0.05 & 0.07 & -0.04 & 0.10 & 0.01 & 0.08 & -0.05 & 0.04 & -0.07 \\
    & $\olsBetai{3}$ & -0.02 & -0.07 & 0.08 & -0.03 & 0.10 & 0.01 & 0.08 & -0.03 & 0.04 & -0.05 \\
    & $\olsBetai{4}$ & -0.02 & -0.07 & 0.07 & -0.03 & 0.09 & 0.01 & 0.08 & -0.03 & 0.04 & -0.05 \\
    & $\olsBetai{5}$ & -0.03 & -0.09 & 0.07 & -0.04 & 0.09 & -0.00 & 0.08 & -0.04 & 0.03 & -0.06 \\
    \bottomrule
  \end{tabular*}
  \caption[Exp. 1: OLS Results. Mixed-Language (Zh), Calibrated.]{Comparison between attention-calibrated (c.) and uncalibrated (\textit{uc.}) mGTE embeddings. Estimated OLS coefficients of the positional fairness analysis (similarity between document embedding and standalone segment embeddings) in the mixed-language document setting (\textbf{Chinese}; $\langRemainder=zh$). $n$ denotes the number of segments per document. OLS regression uses positions as categorical variables; $\olsBetai{0}$ captures the baseline (similarity between document embedding and standalone embedding at position~1), $\olsBetai{p\ge2}$ captures the difference between position $p$'s similarity (similarity between document embedding and standalone embedding at position~$p$) and the baseline similarity. Calibrated embeddings use the following hyperparameters: $\basketSize=128$, $\layersCal=\{7,\ldots,12\}$.}
  \label{tab:ols-coef-multi-zh-mgte-calibrated-comparison}
\end{table*}

\begin{figure*}[tbp]
  \centering
  \includegraphics[width=\textwidth]{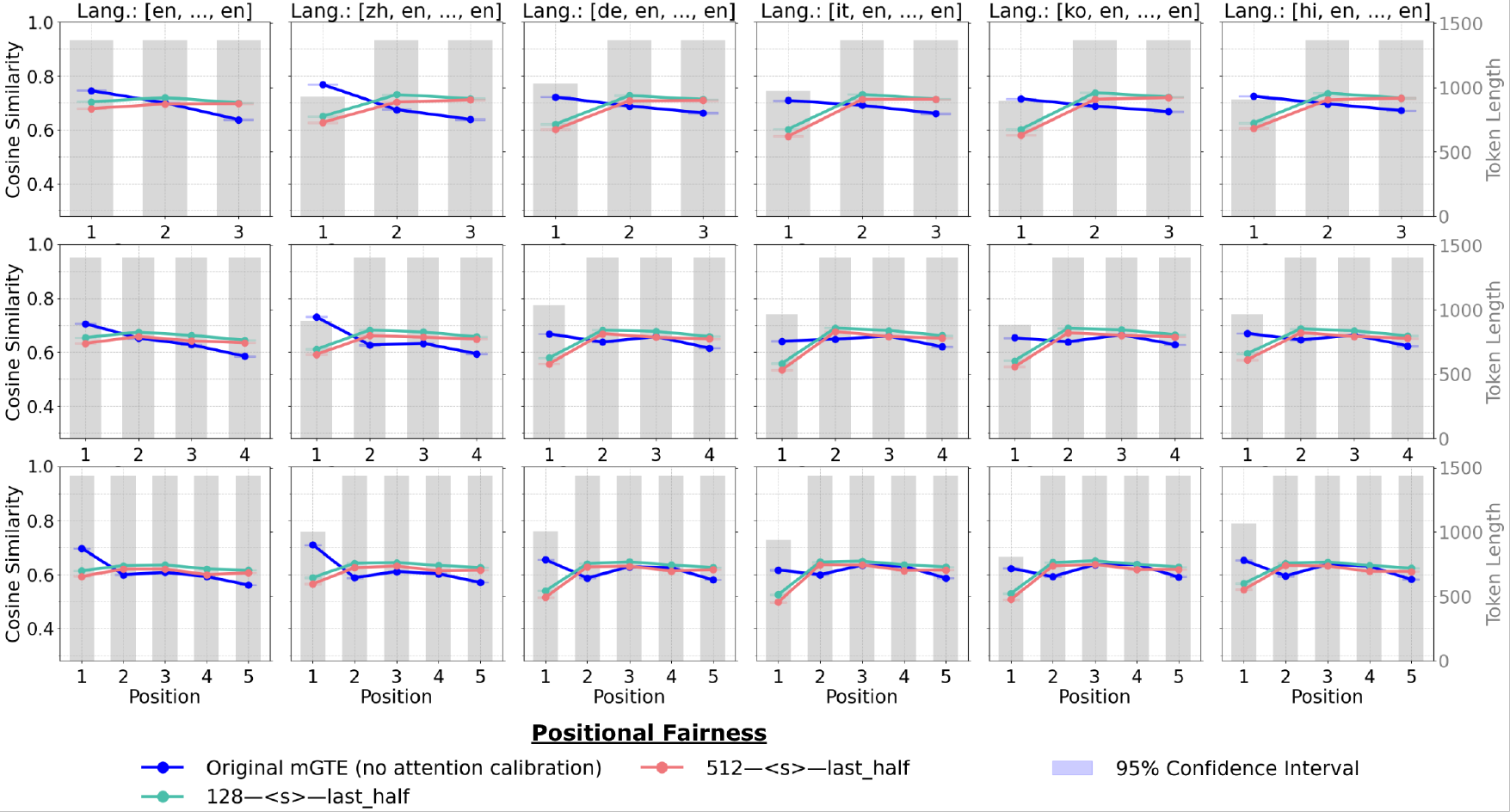}
  \caption{Comparison between attention-calibrated and uncalibrated mGTE embeddings. Subplots show different \textbf{English} ($\langRemainder=en$) mixed-language experiment instances $(n, \langConfig)$, where $n$ varies across rows, and $\langConfig$ varies across columns. Left y-axes show average representation in the global document embedding per segment position. Right y-axes show average token length per segment position (\textcolor{Snow4}{gray bars}). We show two differently parameterized attention calibrations: \textit{128---$<$s$>$---last\_half}: $\basketSize{=}128$, $\layersCal{=}\{7,\ldots,12\}$; \textit{512---$<$s$>$---last\_half}: $\basketSize{=}512$, $\layersCal{=}\{7,\ldots,12\}$.}
  \label{fig:EF_multi_EN_calibrated}
\end{figure*}

\begin{figure*}[tbp]
  \centering
  \includegraphics[width=\textwidth]{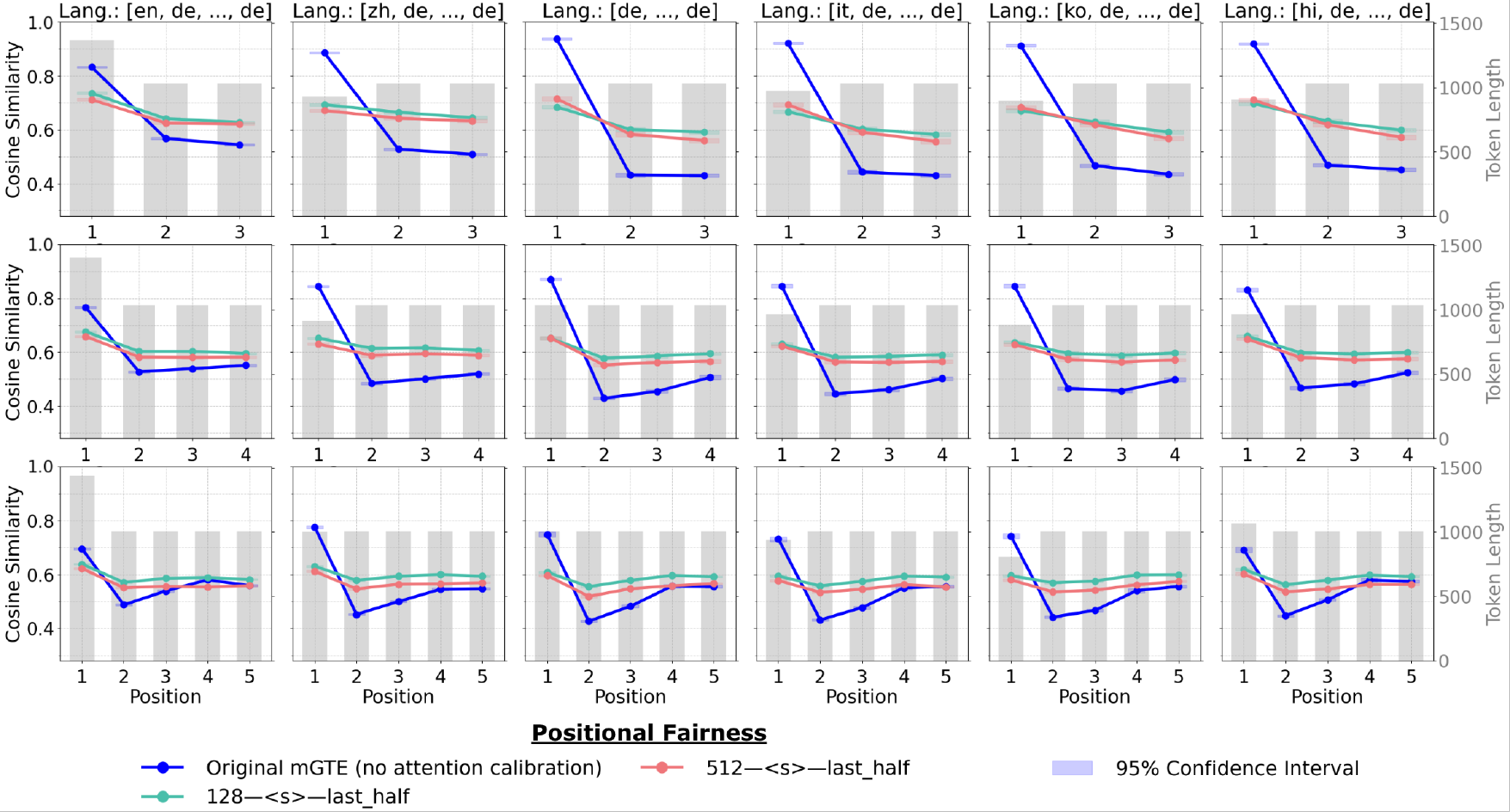}
  \caption{Comparison between attention-calibrated and uncalibrated mGTE embeddings. Subplots show different \textbf{German} ($\langRemainder=de$) mixed-language experiment instances $(n, \langConfig)$, where $n$ varies across rows, and $\langConfig$ varies across columns. Left y-axes show average representation in the global document embedding per segment position. Right y-axes show average token length per segment position (\textcolor{Snow4}{gray bars}). We show two differently parameterized attention calibrations: \textit{128---$<$s$>$---last\_half}: $\basketSize{=}128$, $\layersCal{=}\{7,\ldots,12\}$; \textit{512---$<$s$>$---last\_half}: $\basketSize{=}512$, $\layersCal{=}\{7,\ldots,12\}$.}
  \label{fig:EF_multi_DE_calibrated}
\end{figure*}

\begin{figure*}[tbp]
  \centering
  \includegraphics[width=\textwidth]{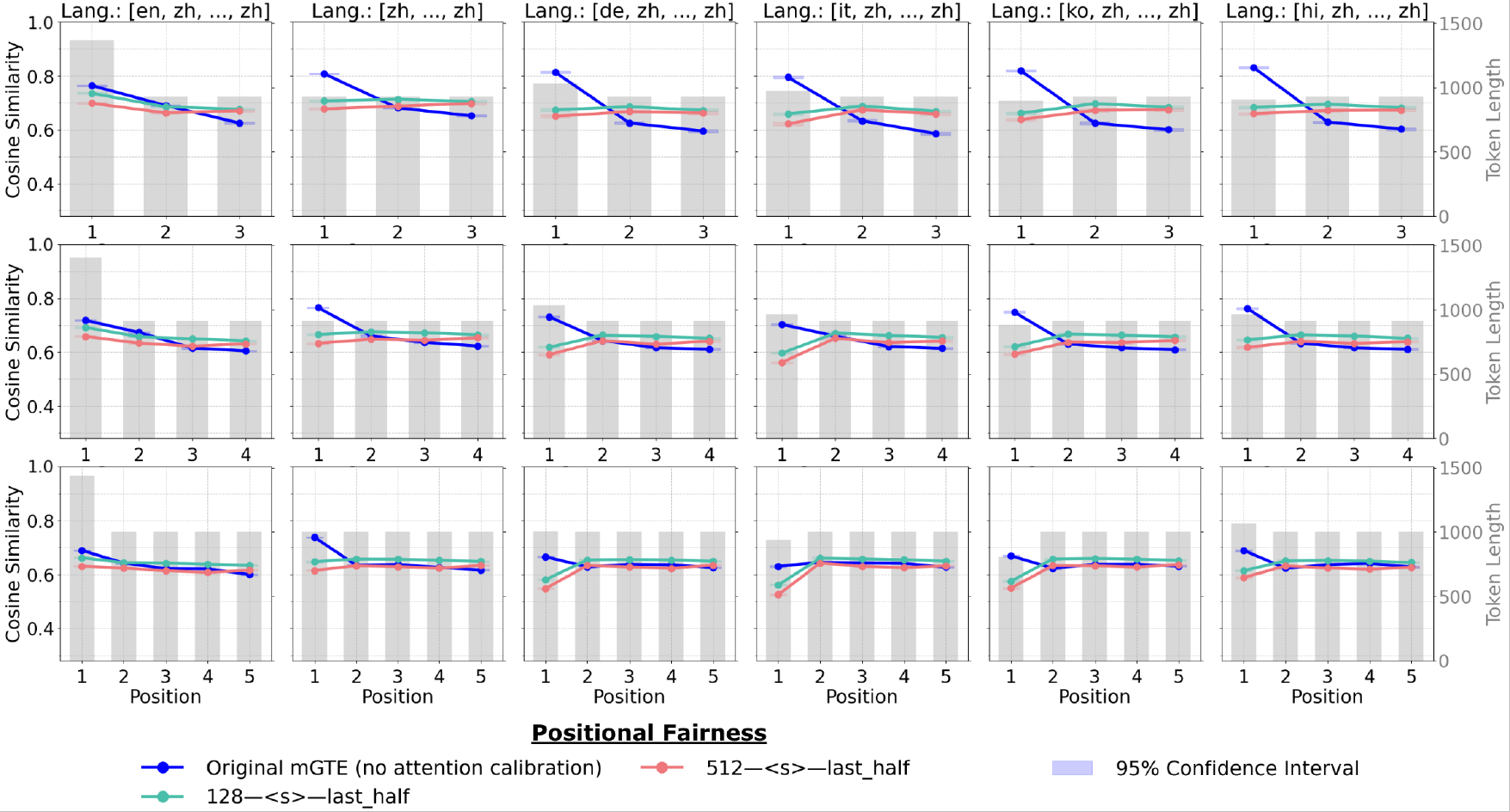}
  \caption{Comparison between attention-calibrated and uncalibrated mGTE embeddings. Subplots show different \textbf{Chinese} ($\langRemainder=zh$) mixed-language experiment instances $(n, \langConfig)$, where $n$ varies across rows, and $\langConfig$ varies across columns. Left y-axes show average representation in the global document embedding per segment position. Right y-axes show average token length per segment position (\textcolor{Snow4}{gray bars}). We show two differently parameterized attention calibrations: \textit{128---$<$s$>$---last\_half}: $\basketSize{=}128$, $\layersCal{=}\{7,\ldots,12\}$; \textit{512---$<$s$>$---last\_half}: $\basketSize{=}512$, $\layersCal{=}\{7,\ldots,12\}$.}
  \label{fig:EF_multi_ZH_calibrated}
\end{figure*}

\end{document}